\pdfoutput=1

\documentclass[11pt]{article}

\usepackage{authblk}
\usepackage[]{EMNLP2023}

\usepackage{times}
\usepackage{latexsym}

\usepackage[T1]{fontenc}
\usepackage{comment}
\usepackage[utf8]{inputenc}

\usepackage{microtype}

\usepackage{inconsolata}

\usepackage{graphicx}
\usepackage{float}
\usepackage{enumitem}
\usepackage{verbatim}
\usepackage{hyperref}
\usepackage{blindtext}
\usepackage{booktabs}
\usepackage{multirow}
\usepackage{array}
\usepackage{subcaption}
\usepackage{amsmath}
\usepackage{paralist}

\usepackage{vcell}

\makeatletter
\renewcommand\AB@affilsepx{\protect\Affilfont}
\makeatother

\title{IndoToxic2024: A Demographically-Enriched Dataset of Hate Speech and Toxicity Types for Indonesian Language}

\author{Lucky Susanto\textsuperscript{*,1}, Musa Izzanardi Wijanarko\textsuperscript{*,1}, Prasetia Anugrah Pratama\textsuperscript{2} \\ Traci Hong\textsuperscript{3}, Ika Idris\textsuperscript{1}, Alham Fikri Aji\textsuperscript{1, 4}, Derry Wijaya}

\affil[*]{Equal Contribution \protect\\}
\affil[1]{Monash University, }
\affil[2]{Independent Researcher, }
\affil[3]{Boston University, }
\affil[4]{MBZUAI}

\begin{document}
\maketitle


\begin{abstract}

Hate speech poses a significant threat to social harmony. Over the past two years, Indonesia has seen a ten-fold increase in the online hate speech ratio, underscoring the urgent need for effective detection mechanisms. However, progress is hindered by the limited availability of labeled data for Indonesian texts. The condition is even worse for marginalized minorities, such as Shia, LGBTQ, and other ethnic minorities because hate speech is underreported and less understood by detection tools. Furthermore, the lack of accommodation for subjectivity in current datasets compounds this issue. To address this, we introduce IndoToxic2024, a comprehensive Indonesian hate speech and toxicity classification dataset. Comprising 43,692 entries annotated by 19 diverse individuals, the dataset focuses on texts targeting vulnerable groups in Indonesia, specifically during the hottest political event in the country: the presidential election. We establish baselines for seven binary classification tasks, achieving a macro-F1 score of 0.78 with a BERT model (IndoBERTweet) fine-tuned for hate speech classification. Furthermore, we demonstrate how incorporating demographic information can enhance the zero-shot performance of the large language model, gpt-3.5-turbo. However, we also caution that an overemphasis on demographic information can negatively impact the fine-tuned model performance due to data fragmentation.

\end{abstract}

\section{Introduction}
\begin{figure}[t]
    \centering
    \includegraphics[width=0.4\textwidth]{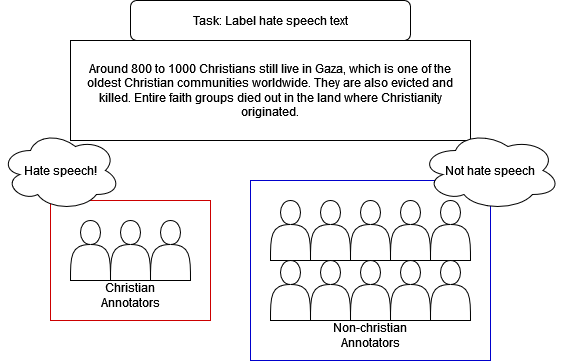}
    \caption{The perception of hate speech is influenced by the identity of a person or a group. A text may be considered hate speech by one group of people, while another group may not view it as such. More divisive example available in appendix \ref{app:divisive-text}.}  
    \label{fig:demographic-subjectivity}
    \vspace{-15pt}
\end{figure}

In the rapidly evolving digital landscape of Indonesia, a disturbing \textbf{ten-fold increase in hate speech} ratio has been observed in just two years \cite{csis, aji-hatespeech-dashboard}. Left alone, this surge threatens social harmony \cite{hatespeech-williams-2019}, and is especially harmful to minority groups \cite{alexandra-2023},  because it could lead to societal polarization \cite{unlu-2024}. One potential solution comes in the form of an automated hate speech detection system. However, many significant challenges to the development of this system exist. One of these challenges is the lack of comprehensive and up-to-date data. The existing Indonesian datasets \citep{alfina2017,ibrohim2018} are considerably dated and consist only of around 3,000 labeled Indonesian texts. 
Furthermore, these datasets lack crucial information, such as the demographic information of annotators, which is key in training systems since subjectivity is inherent in hate speech annotations \cite{majority-fleisig-2024} (Figure \ref{fig:demographic-subjectivity}).
Subjectivity in the annotation of text often occurs with latent content (e.g., sarcasm) which contains "under the surface" information that requires human annotators' mental scheme to decipher their meaning. In contrast, manifest content (e.g., how many words are in the sentence) is "surface information" that requires minimal interpretation \cite{lombard2002content}. Arguably, the more pertinent information is found in latent content. However, human annotator's judgments are affected by not only their background, geography, and personal experiences \cite{armstrong1997place} but also by their perceived status in relation to other human coders working on the research \cite{campbell2013coding}.

As a first step to tackle this issue, we release \textbf{IndoToxic2024}\footnote{The code is available at \url{https://github.com/izzako/IndoToxic2024/tree/main}}, a hate speech dataset annotated across various demographics, where each entry is accompanied by ten-dimensional demographic information. Unlike most hate speech datasets that are single-labeled, IndoToxic2024's labels are preserved per annotator. This opens up the potential for studying subjectivity in annotations. This dataset consists of 43,692 entries annotated by 19 diverse annotators, aimed at classifying hate speech and the types of toxic behaviors. It is a human-annotated collection, assembled by gathering posts from various social media platforms using keywords related to Indonesian vulnerable groups. Our contributions are three-fold:

\begin{compactitem}
    \item \textbf{Creation of IndoToxic2024 Dataset,} enabling the creation of better hate speech detection systems (IndoBERTweet fine-tuned for hate speech classification), specifically in the Indonesian language. 
    \item \textbf{Exploring the Role of Demographic Information in Hate Speech Classification,} demonstrating that gpt-3.5-turbo’s hate speech classification performance can be significantly improved when provided with the demographic information of the annotator. This insight suggests that demographic information can be a valuable addition to improve model performance. 
    \item \textbf{Analyzing the Impact of Excessive Demographic Information on Fine-Tuned Model Performance,} providing an extensive analysis that shows an addition of demographic information can lead to fragmentation of the dataset (fewer training data in each demographic, thus worse performance). 
\end{compactitem}


\section{Background \& Related Work}


\subsection{What is Hate Speech?} 
Hate speech definitions evolved over time. Initially defined as language intended to demean others \cite{hsdef-delgado-1982}, it has expanded to include public speech or writing inciting hatred towards demographic groups \cite{hsdef-greenawalt-1989, hsdef-un-2023}. In this work, we adopt the definition by Indonesia’s National Human Rights Commission, which includes any communication motivated by hatred against people based on their identities, intending to incite violence, death, and social unrest \cite{hsdef-paramadina-2023}.

\subsection{Detecting Hate Speech} 
Deep learning techniques have been promising in automatic hate speech detection \citep{hsclassifier-jigsaw,hsclassifier-multilang-das-2021}.
However, these techniques still fall short in real-world scenarios due to the evolving nature of hate speech and the complexity of the task. Recently, \citet{majority-fleisig-2024} demonstrated that incorporating demographic information of annotators can improve model performance. \citet{majority-fleisig-2024} approaches the problem as a regression instead of a classification, where texts are labeled based on the severity of their toxicity. 

\subsection{Available Hate Speech Datasets}
Initially, Indonesian hate speech datasets \citep{alfina2017}, consisted of texts with a binary label indicating is a hate speech or not. As the field evolved, datasets began to incorporate different levels of toxicity. For instance, the datasets by \citet{ibrohim2018} and \citet{mathew2022hatexplain} introduced varying degrees of toxicity, with the latter focusing on explainable hate speech classification. Following this trend, datasets started to include types of hate speech, as seen in the challenge run by \citet{hsclassifier-jigsaw}. More recently, demographic information has been incorporated into hate speech datasets. To our knowledge, the earliest dataset to contain more than just text and labels was created by \citet{kumar-2021} and was recently utilized by \citet{majority-fleisig-2024}. Altough, it’s important to note that most modern datasets focus on the English language. Aside from the private \citet{csis} dataset, we have not found new Indonesian hate speech datasets in the past five years.

\subsection{Datasets With Demographic Information} 
Datasets incorporating demographic information are typically utilized for tasks associated with an individual’s behavior or circumstances. More often than not, this demographic data is employed in predicting aspects such as insurance premiums \cite{medical-insurance} or an individual’s purchasing power \cite{consumer-behaviour}. However, considering the subjectivity inherent in hate speech, we only find a single hate speech dataset that includes demographic information \cite{kumar-2021}. Most hate speech datasets only go as far as providing annotator IDs, as in the case of \citet{mathew2022hatexplain}, without any demographic information of the text annotator.
\section{Dataset Creation}

\subsection{Data Collection}
We obtain our text data from some of the popular social media platforms in Indonesia \citep{kemp-2023} including Facebook, Instagram, and Twitter (or X). Additionally, we also retrieve articles from CekFakta\footnote{\url{https://cekfakta.com/}}, a movement that focuses on clarifying misinformation that spreads on the internet.

Different tools are utilized to gather data from platforms. We use \citet{brandwatch} to obtain tweets, replies, and quotes from X, Crowdtangle \cite{crowdtangle} to obtain posts from Instagram and Facebook. Additionally, we retrieved articles from Cek Fakta\footnote{\url{https://mafindo.or.id}}, a fact-checking collaboration across online media and fact-checking organizations in the country. 
We collect data from September 2023 to January 2024, following the 2024 Indonesian presidential election timeline 
\citep{detikcom-2022}, as hate speech was found to intensify in Indonesia during a similar election in 2019 \cite{iswatiningsih2019hate}.

We obtain data using keywords that were previously used to express hate toward vulnerable groups in Indonesian texts, compiled 
based on various sources such as literature research, discussions with experts, and a focus group discussion (FGD)  with representatives from vulnerable communities\footnote{identified minority groups in Indonesia that have been the target of hate speech in previous elections including disability, LGBTQ+, Chinese, Ahmadiyya, Shia, Catholics, and Christian groups.}. We provide the keywords in Appendix \ref{sec:scrape-keywords}. The data was then sampled for annotation by \textit{coders} (i.e. annotators from diverse demographic backgrounds).

\subsection{Recruitment and Validation Metrics}
18 people from various demographic backgrounds and 1 from our research team were recruited to annotate the data. From the FGD, each of the vulnerable groups proposed their representatives to annotate the data, ensuring representations from each group. Table \ref{tab:annotdemo} gives us a coarse-grained overview of this diversity. 
Subsequently, contracts were drafted, and each annotator was compensated with 1.5 million IDR for every 1,000 texts they annotated. For comparison, an average monthly wage in Indonesia across sectors is 3.5 million IDR in 2024  \cite{bpsAverageWageSalary}. 

\begin{table}[ht]
\small
\centering
\resizebox{\columnwidth}{!}{
\begin{tabular}[ht]{lrr}
\toprule
\textbf{Demographic} & \textbf{Group}& \textbf{Count}\\
\midrule
Disability  & \begin{tabular}[t]{@{}r@{}}With Disability\\ No Disability\end{tabular}  & \begin{tabular}[t]{@{}r@{}}3\\ 16\end{tabular}\\[0.25cm] 
 \hline
 Ethnicity  & \begin{tabular}[t]{@{}r@{}}Chinese\\ Indigeneous\\Other\end{tabular}  & \begin{tabular}[t]{@{}r@{}}3\\ 15\\1\end{tabular}\\[0.25cm] 
 \hline
Religion & \begin{tabular}[t]{@{}r@{}}Islam\\ Christian or Catholics\\ Hinduism or Buddhism\\ Ahmadiyya or Shia\\ Traditional Beliefs\end{tabular} & \begin{tabular}[t]{@{}r@{}}9\\ 4\\ 3\\ 2\\ 1\end{tabular}\\
\hline
Gender  & \begin{tabular}[t]{@{}r@{}}Male\\ Female\end{tabular}                                       & \begin{tabular}[t]{@{}r@{}}6\\ 13\end{tabular}\\ 
 \hline
LGBTQ+  & \begin{tabular}[t]{@{}r@{}}Yes\\ No\end{tabular}                                       & \begin{tabular}[t]{@{}r@{}}1\\ 18\end{tabular}\\ 
\hline
Age  & \begin{tabular}[t]{@{}r@{}}18 - 24\\ 25 - 34\\ 35 - 44\\ 45 - 54\end{tabular}   & \begin{tabular}[t]{@{}r@{}}9\\ 5\\ 3\\ 2\end{tabular}\\ 
\hline
Education & \begin{tabular}[t]{@{}r@{}}Master's Degree\\ Bachelor's Degree\\ Associate's Degree\\ High School Degree\end{tabular}                 & \begin{tabular}[t]{@{}r@{}}3\\ 7\\ 2\\ 7\end{tabular}\\
\hline
Job Status   & \begin{tabular}[t]{@{}r@{}}Employed\\ College Student\\ Housewife or Unemployed\end{tabular}                                           & \begin{tabular}[t]{@{}r@{}}9\\ 8\\ 2\end{tabular} \\ 
\hline
Presidential Vote &\begin{tabular}[t]{@{}r@{}}Candidate no. 1\\ Candidate no. 2\\ Candidate no. 3\\ Unknown or Abstain\end{tabular}                                             & \begin{tabular}[t]{@{}r@{}}4\\ 7\\ 5\\3\end{tabular} \\
\bottomrule
\end{tabular}
}
\caption{The demographic background of the 19 annotators in coarser-granularity. The ethnicity demographic information that we have are more fine-grained where \textit{Indigenous} group here refers to several ethnic Indonesian groups: Java, Minang, Sunda, Bali, Dayak, Bugis, etc. with 1-2 annotators per ethnicity.}
\label{tab:annotdemo}
\vspace{-15pt}
\end{table}

\textbf{Inter-coder Reliability Metrics} \quad
Coders were trained on a codebook and their agreement on applying the codebook on the data was determined by calculating inter-coder reliability (ICR). High ICR score means that the annotators consistently categorized the text similarly. Although Cohen's Kappa is a popular inter-coder reliability test used in several hate speech works 
\cite{Aldreabi2023,ayele-etal-2023-exploring,vo2024exploiting}, Gwet's AC1 has been suggested as a more stable metric that is robust against class imbalance \citep{Ohyama2021,Wongpakaran2013}. This is especially relevant for social media platforms with a high volume of data where the majority is non-hate speech. 

\subsection{Annotation Instrument}
Our goal is to capture text that contains toxicity, be it explicit (manifest content such as inclusion of offensive words) or implicit (latent content such as sarcasm) \cite{krippendorff2018content}. This nuanced and contextual hate speech has not yet been confronted in a consistent and unified manner in the NLP community \cite{elsherief-etal-2021-latent}.

Based on the literature review of hate speech and discussions with vulnerable groups in Indonesia, we create a codebook (Appendix \ref{sec:annotguide}) as 
a guide for annotators to identify text as hate speech when certain criteria are met \cite[p.25-30]{Sellars_2016}. The codebook helps annotators recognize text typically seen as hate speech and text that seems normal but is indeed harmful to a specific vulnerable group. 

\subsection{Annotation Process}
The annotation process is divided into two stages: the training phase and the main annotation phase. The annotators were trained on the codebook during the training phase. They were instructed to 
identify whether the text contains hate speech (or toxicity). If yes, they were asked to identify the types of the hate speech (i.e., whether it is an insult, threat, profanity, identity attack, or sexually explicit text). 

During the first training session, the annotators were given 100 randomly-sampled text to code (i.e., annotate), but ICR was not met. Hence, a second (with another 100 randomly-sampled text) and a third (with another 249 randomly-sampled text) training session were held to further clarify the codebook and resolve any confusion. 
A satisfactory ICR score was met after the third training session: a Gwet's AC1 score of $0.61$ for the toxicity label. Following Quantitative Content Analysis (QCA) in communication research, which is a commonly used method to derive replicable and meaningful inferences from texts \cite{krippendorff2018content}, once the ICR was met, the annotators continued to code more texts independently in the main annotation phase. 
Our final data is comprised of 43,692 texts that were annotated from the three sessions of training and the main annotation phase\footnote{Due to the long annotation process and the complexity and human toll of the task, some annotators complete only parts of their assignments during the main annotation phase.}. 
\vspace{-6pt}
\subsection{Dataset Properties}
Out of 43,692 texts in IndoToxic2024, 6,894 are labeled as toxic. 
From Table \ref{tab:toxictype-labelcount}, we can observe that almost half of the toxic texts are insults.

\begin{table}[h]
\centering
\small
\begin{tabular}{lc}
\hline
\textbf{Toxicity Types} & \textbf{\# Texts} \\
\hline
Insults & 3140 \\
Threat / Incitement to Violence & 2837 \\
Profanity or Obscenity & 1271 \\
Identity Attack & 1061 \\
Sexually Explicit & 224 \\
\hline
\end{tabular}
\caption{Number of toxic texts labeled with types. These categories are not mutually exclusive, since a toxic comment could exhibit more than one type of hate speech.}
\label{tab:toxictype-labelcount}
\vspace{-10pt}
\end{table}


We can further explore how different demographic groups annotate the texts. When we aggregate the dataset, we see that the distribution of toxicity labels differs between genders. Males labeled 19.3\% of their data as toxic, whereas females labeled 13\%. We explore this topic in more depth in the next section.
\section{Dataset Analysis: Existence of Subjectivity}
\begin{figure*}[t]
    \centering
    \includegraphics[width=\textwidth]{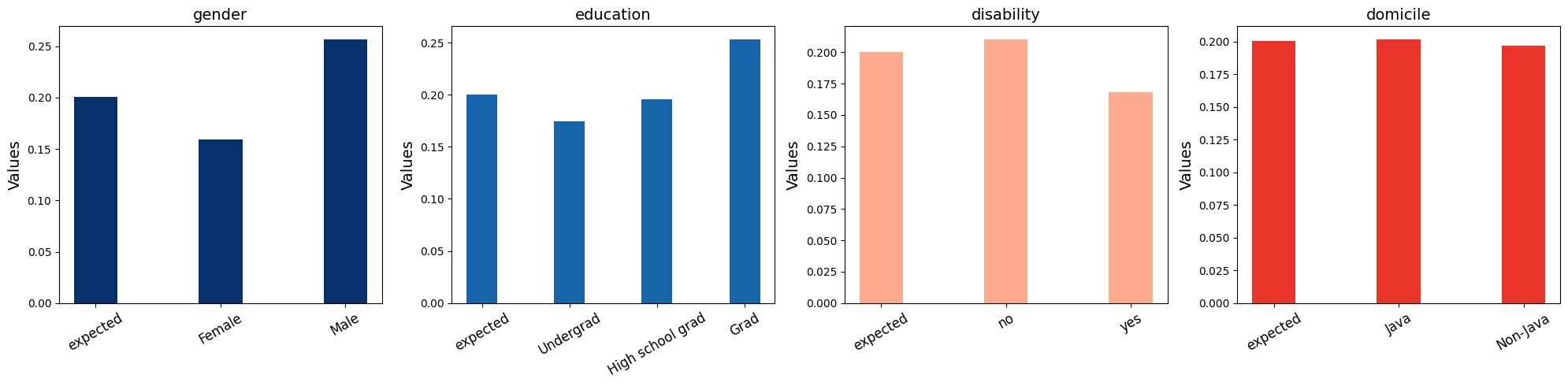}
    \vspace{-20pt}
    \caption{\textbf{Subjectivity affects the annotation process.} Expected ratio of hate speech vs. non-hate speech labeled data against the ratio of each demographic group's annotations using chi-square testing. Blue colors indicate a lower p-value (more significant difference to the expected ratio), and red colors indicate a higher p-value (less significant difference to the expected ratio).}  
    \label{fig:subjectivity-distribution}
    \vspace{-15pt}
\end{figure*}
\subsection{Subjectivity in Hate Speech Annotations} 
The research by \citet{majority-fleisig-2024} indicates that combining demographic data with potential hate speech improves toxicity prediction models. \citet{kumar-2021} further emphasizes the subjective nature of toxic text, as shown by annotator disagreements. Our study expands on these insights by exploring the topic from three new perspectives.

\paragraph{Through Distribution Assumptions} 
In the annotation process, texts are randomly assigned without considering the annotator’s gender, education, disability, or domicile. We use chi-square testing to verify the null hypothesis: “If texts are randomly assigned and there’s no subjectivity, each annotator should receive a similar proportion of hate speech texts”. However, Figure \ref{fig:subjectivity-distribution} shows that the null hypothesis is only valid for the “domicile” group. Therefore, we reject the null hypothesis, confirming subjectivity in most demographic groups. It’s important to note that some texts were assigned based on the religion or ethnicity of the annotator, so these groups are excluded from this analysis.

\paragraph{Through ICR Score} 
We calculate the ICR score within each demographic group (within-group ICR score) and between two demographic groups (between-group ICR score) (refer to Appendix \ref{sec:rigorICR} for more details). It’s generally assumed that the between-group ICR scores would be lower than the within-group ICR scores due to subjectivity within the group. However, our results mostly contradict this assumption. For example, while the within-group ICR score for females is 0.61 and for males is 0.54, the between-group ICR score is 0.58, which is not lower than both within-group scores. Further analysis reveals the role of intersectional identity. For instance, a high ICR score contributor among the female-male pairs belongs to other similar demographic groups (both are disabled annotators and identify as Christians). This suggests that demographics are intersecting factors, not mutually exclusive aspects.


\vspace{-6pt}
\paragraph{Through Modeling Results} 
Fine-tuning a model using a subjective or biased dataset can result in a model that inherits that subjectivity \citep{sengupta2022causal}. To ensure the availability and quantity of the training dataset, we use a coarser granularity for some of our 
demographic categories. For instance, we use age groups instead of specific age numbers for the age demography, and 
a coarser grouping of Islam and Non-Islam for the religious demography. We then fine-tune IndoBERTweet model \cite{indobert-koto-2021} by limiting the training data to texts that are annotated by the same group (e.g., Islam), 
and test on texts that are annotated by the other group in the demographic category (e.g., Non-Islam). As a result, a model fine-tuned on \textit{Non-Islam} annotators' labels tends to perform worse when tested on \textit{Islam} annotators' labels compared to when tested on their own labels (using 5-fold cross-validation) and vice versa (see Table \ref{tab:perf-test-on-other-group}). This trend holds true for other demographic categories such as disability, ethnicity, and religion. 
The result indicates that for some demographic groups, there may exist other demographic groups that annotate the texts differently due to differences in their identities.

\begin{table}
\centering
\small
\begin{tabular}{>{\hspace{0pt}}p{0.335\linewidth}>{\centering\hspace{0pt}}m{0.335\linewidth}>{\centering\arraybackslash\hspace{0pt}}m{0.19\linewidth}} 
\toprule
\multirow{2}{\linewidth}{\hspace{0pt}\textbf{Train on}}  & \multicolumn{2}{>{\centering\arraybackslash\hspace{0pt}}p{0.525\linewidth}}{\vcell{\textbf{Test on}}}                     \\[-\rowheight]
  & \multicolumn{2}{>{\centering\arraybackslash\hspace{0pt}}m{0.525\linewidth}}{\printcellmiddle}                             \\ 
\cline{2-3}
 & \multicolumn{1}{>{\hspace{0pt}}m{0.335\linewidth}}{Non-Islam} & \multicolumn{1}{>{\hspace{0pt}}m{0.19\linewidth}}{Islam}  \\ 
\midrule
\multicolumn{1}{>{\hspace{0pt}}m{0.335\linewidth}}{Non-Islam} & 0.605                                                            & 0.597                                                        \\
\multicolumn{1}{>{\hspace{0pt}}m{0.335\linewidth}}{Islam}     & 0.628                                                            & 0.66                                                        \\
\bottomrule
\end{tabular}
\caption{\textbf{Subjectivity affects performance of models.} F1-scores of IndoBERTweet, trained on \textit{Islam} and \textit{Non-Islam} annotators' labels, using 5-fold cross-validation when the training and testing data are from the same group's annotations (i.e., the diagonals). It can be seen that the performance of training on one group's annotations and testing on another is worse than training and testing on the same group's annotations.}
\label{tab:perf-test-on-other-group}
\vspace{-15pt}
\end{table}

\subsection{Texts with Differing Annotations}
Accompanying previous results, we rank texts based on their divisiveness. A text is highly divisive when groups of annotators in the same demographic category largely disagree on their annotations. For example, the text in Figure \ref{fig:demographic-subjectivity} where Christian annotators unanimously agree that the text \textit{is} hate speech while the non-Christians unanimously disagree. 
More examples can be found in appendix \ref{app:divisive-text}. 
The text illustrates how the interpretation of hate speech can vary depending on the annotator’s demography, in this case their religion.  However, it is crucial to consider the whole picture. There are instances where a text appears to be on the topic of a specific demographic category, such as female, yet annotators are split not along gender but along their (last) education level. 

\subsection{How Topic Affects Subjectivity}

\begin{figure}[h]
    \centering
    \includegraphics[width=0.5\textwidth]{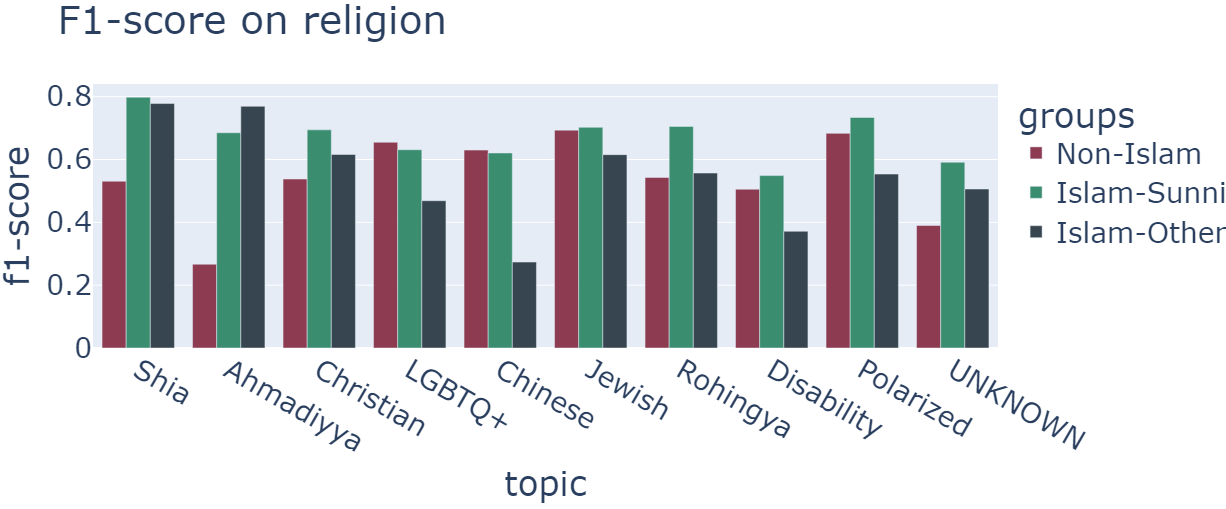}
    \vspace{-15pt}
    \caption{\textbf{Topics can act as a stabilizer or a catalyst.} Some topics can be \textit{easier} for a certain demography to annotate. For example, texts relating to Shia are \textit{easier} for Islamic group of people to annotate while \textit{harder} for non-Islam people to annotate.  
    }  
    \label{fig:subjectivity-and-topic}
    \vspace{-10pt}
\end{figure}

We trained a hate speech classifier on texts annotated by each demographic group and then checked how well each model did on different topics. Topics of a text are target demographics of the text, identified based on keywords mentioned in the text. The keywords-to-topics mapping is in appendix \ref{sec:mapping_topics}. 

An example of how models performed across topics is shown in Figure \ref{fig:subjectivity-and-topic}. 
One model was trained on a dataset annotated by people \textit{with} non-Muslim religion ("Non-Islam"), one was trained on a dataset annotated by people \textit{with} Islam Sunni religion ("Islam-Sunni"), and the third one trained on a dataset annotated by people \textit{with} "Shia" or "Ahmadiyya" as their religion ("Islam-Other"). For topics relating to Shia or Ahmadiyya, both of the "Islam" models perform similarly while the model trained on "Non-Islam" dataset perform relative poorly. However, for topics relating to 
LGBTQ+, or Chinese, there’s a jump in performance for the "Non-Islam" model. 
This suggests that there’s less disagreement among people of the same group when they understand or relate to the topics, and higher disagreement when they are unfamiliar with the topics.
\section{Benchmark Results, Experiments, and Analysis}

\subsection{Baseline Model Performance per Task}
\begin{table*}[htb]
    \small
    \centering
    \begin{tabular}{lcccccccccc}
        \toprule
        & \multicolumn{4}{c}{\textbf{Metrics}} & \multicolumn{5}{c}{\textbf{Data Statistic}} \\
        \cmidrule(lr){2-5} \cmidrule(lr){6-10}
        \textbf{Classification Task} & \textbf{Accuracy} & \textbf{F1} & \textbf{Precision} & \textbf{Recall} & \textbf{Our-0} & \textbf{Our-1} & \textbf{CSIS-0} & \textbf{CSIS-1} & \textbf{Synthetic-1} \\
        \midrule
        Hate Speech                 & 0.89       & 0.78     & 0.80       & 0.76     & 4014 & 1338 & 21125 & 7112    & 1325 \\
        Identity Attack         	& 0.75       & 0.80     & 0.74       & 0.88     & 461  & 648  & -     & -       & 228 \\
        Incitement to Violence      & 0.77       & 0.53     & 0.55       & 0.52     & 345  & 115  & 945   & 315     & 890 \\ 
        Insult                      & 0.79       & 0.85     & 0.81       & 0.88     & 705  & 423  & 149   & 1142    & 785 \\ 
        Profanity or Obscenity  	& 0.81       & 0.70     & 0.70       & 0.72     & 824  & 373  & -     & -       & 370 \\ 
        Sexual Explicit         	& 0.91       & 0.80     & 0.88       & 0.77     & 123  & 41   & -     & -       & 82 \\
        \bottomrule
    \end{tabular}
    \caption{Performance of IndoBERTweet across various binary classification tasks, utilizing a combination of our annotated data, data from CSIS, and synthetically generated data via GPT-3.5-turbo. The term \textbf{-x} represents the quantity of data associated with that label for a specific task. Our sampling strategy ensures that the quantity of 0-labeled (e.g., non-hate speech) data is at most three times that of 1-labeled  (e.g., hate speech) data. 
    }
    \label{tab:perf-and-data}
    \vspace{-15pt}
\end{table*}
We benchmark 6 classification tasks to identify: 
whether a text contains hate speech, and the five types of hate speech. We fine-tune IndoBERTweet \cite{indobert-koto-2021} on our dataset. To enhance the performance of our baselines, we merged our dataset with that of \citet{csis} and generated synthetic hate speech data using gpt-3.5-turbo \cite{brown2020language} through 10-shot generation. 
This synthetic data was only used for training. 
The stratified 10-fold performance and data breakdown for each task are reported in Table \ref{tab:perf-and-data}. The prompt given to gpt-3.5-turbo to generate synthetic texts is available in appendix \ref{app:synthetic-gen-prompt}.

For most tasks, our baseline achieved a macro F1-score above 0.7. However, for the "incitement to violence" classification task, we only achieved a macro F1-score of 0.53. It’s important to note that the data reported in Table \ref{tab:perf-and-data} were used to achieve the best baseline performance reported here. 

\subsection{Incorporating Demographic and Topic Information}
\citet{majority-fleisig-2024} shows that incorporating metadata such as demographic information and survey results can increase a hate speech classifier's performance. For the following set of experiments, we report on the performance of three models: IndoBERTweet \cite{indobert-koto-2021}, gpt-3.5-turbo \cite{brown2020language}, and SeaLLM-7B-v2.5 \cite{nguyen2023seallms}. IndoBERTweet and SeaLLM are pre-trained with a focus on Indonesia and SouthEast Asian (SEA) languages respectively.  

Without any demographic or topic information, IndoBERTweet, fine-tuned only on IndoToxic2024, has the best performance for hate speech classification with a macro-F1 of 0.718 (5-fold cross-validation). We also report the zero-shot performance of the other models in Table \ref{tab:concise-baseline}a. 
The performance of IndoBERTweet, after incorporating topic and/or demographic information, is detailed in Table \ref{tab:concise-baseline}b.

\begin{table}[t]
    \centering
    \begin{subtable}[b]{0.5\linewidth}
        \centering
        \resizebox{\linewidth}{!}{
        \begin{tabular}{l p{1.3cm}}
          \toprule
          \textbf{Model} & \textbf{F1} \\
          \toprule
            \textbf{IndoBERTweet} & \textbf{0.718} \\
            GPT-3.5-turbo & 0.627 \\
            SeaLLM-7B-v2.5 & 0.517 \\
          \bottomrule
        \end{tabular}
        }
        \caption{Baseline performance}
        \label{tab:natural-generative-results-by-models}
    \end{subtable}
    \hfill
    \begin{subtable}[b]{0.48\linewidth}
    \centering
        \resizebox{\linewidth}{!}{
        \begin{tabular}{p{2.8cm} p{1.3cm}}
          \toprule
          \textbf{Information} & \textbf{F1} \\
          \toprule
            Baseline & 0.718 \\
             + Demographic & 0.672 \\
             \textbf{+ Topic} & \textbf{0.755} \\
             + Topic \& Demo & 0.709 \\
          \bottomrule
        \end{tabular}
        }
        \caption{IndoBERTweet performance with augmented input.}
        \label{tab:natural-generative-results-by-langs}
    \end{subtable}
    \caption{IndoBERTweet, when given only topic information, performs best based on the macro-F1 metric. Models are trained using only \textbf{IndoToxic2024} dataset.}
    \label{tab:concise-baseline}
    \vspace{-15pt}
\end{table}


To incorporate the topic (\verb|t|) and demographic information (\verb|d|) along with the texts (\verb|w|), the input is formatted as follows:
\[
d_1 \ldots d_n \, \,t_1 \ldots t_n \, [SEP] \, w_1 \ldots w_n
\]
For instance, a complete input given to IndoBERTweet might be: \textit{“Reader information: Chinese ethnicity, Christian, Male, Millennial. Topic: Christian. [SEP] [TEXT]”}. 

To use gpt-3.5-turbo and SeaLLM-7B-v2.5, a custom prompt is used, with its system prompt set to: \textit{“You may only respond with either a 0 or a 1”}. We also enforce the statement in the user input, which follows this pattern: \textit{“Is the Indonesian text below a hate speech [Demographic Information]? If yes, respond with 1, otherwise respond with 0. [TEXT]”}. 

\begin{figure}[h]
    \centering
    \includegraphics[width=0.5\textwidth]{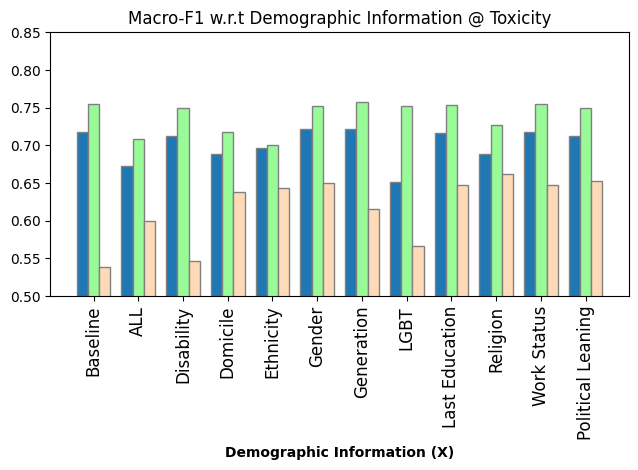}
    \caption{Comparison of macro-F1 Score from IndoBERTweet (dark green/left bars), IndoBERTweet with topic (light green/middle bars), and gpt-3.5-turbo (orange/right bars) given varying degrees of demographic information. Baseline means no demographic information was given.}  
    \label{fig:toxicity-perfs}
    \vspace{-12pt}
\end{figure}

Each text in our dataset is accompanied by demographic information about the annotator. This information includes disability status, domicile, ethnicity, gender, generation (age group), part of the LGBTQ+ communities, last education, religion, work status, and political leaning. Our experiments are divided into three parts: one with no demographic information provided, one with all demographic information provided, and one with only a single piece of demographic information provided, totaling 12 experiments. These experiments are conducted on three models: IndoBERTweet, IndoBERTweet with topic information given, and gpt-3.5-turbo. The performance of these models can be found in Figure \ref{fig:toxicity-perfs}.

Our findings indicate that for IndoBERTweet models, providing demographic information does not yield any significant improvement. In fact, it may even negatively impact the model’s performance. However, in all instances, the performance of gpt-3.5-turbo improved when demographic information was provided. We hypothesize that the additional information adds a level of complexity during the fine-tuning of the IndoBERTweet model that exceeds what the dataset can support.

\subsection{Ablation of Impact per Topic}

\begin{figure}[h]
    \centering
    \includegraphics[width=0.5\textwidth]{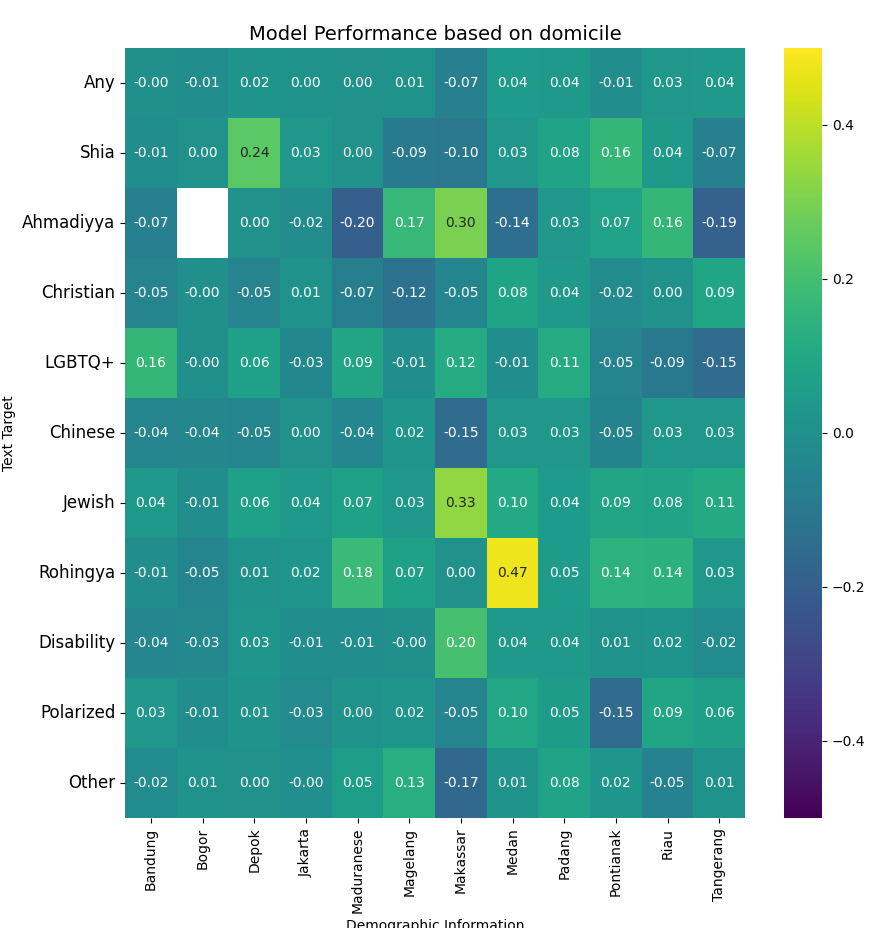}
    \vspace{-15pt}
    \caption{The effect ($\Delta$ F1) of giving domicile information to gpt-3.5-turbo on its hate speech text classification compared to not giving it any information.}  
    \label{fig:gpt-demo-domicile}
    \vspace{-15pt}
\end{figure}

\paragraph{Leveraging its Pre-training, gpt-3.5-turbo Understands Cultural and Value Differences.} As indicated by its performance in Figure \ref{fig:gpt-demo-domicile}, gpt-3.5-turbo, thanks to its pre-training data, inherently understands the cultural and value differences among people based on their identities. The figure shows that gpt-3.5-turbo significantly improves when provided with the annotator’s domicile, without any fine-tuning. For instance, it achieved a 0.47 macro-F1 improvement when the annotator resided in \textbf{Medan} and the text targeted \textbf{Rohingya} refugees. We attribute this improvement to the training data, which might include articles about Rohingya refugees often first arriving in Medan \cite{un-rohingya-medan-2024}. Another example is \textbf{Makassar} and text targeting the \textbf{Disabled}, where the model achieves a 0.2 macro-F1 improvement when given demographic information. This could be due to the model’s understanding of the unique relationship of \textbf{Disabled} people in \textbf{Makassar} compared to other locations \cite{makasar-disabled-post-2023}. Without demographic information, we assume that the model defaults to an inherent bias. For instance, when domicile information is not provided, gpt-3.5-turbo seems to assume the reader’s location is either the Indonesian capital \textbf{Jakarta} or another big city, \textbf{Bogor}, as suggested by the non-significant performance gap visible in Figure \ref{fig:gpt-demo-domicile}. Visualizations for other cases are available in the appendix \ref{app:heatmaps}.

\begin{figure}[h]
    \vspace{-10pt}
    \centering
    \includegraphics[width=0.5\textwidth]{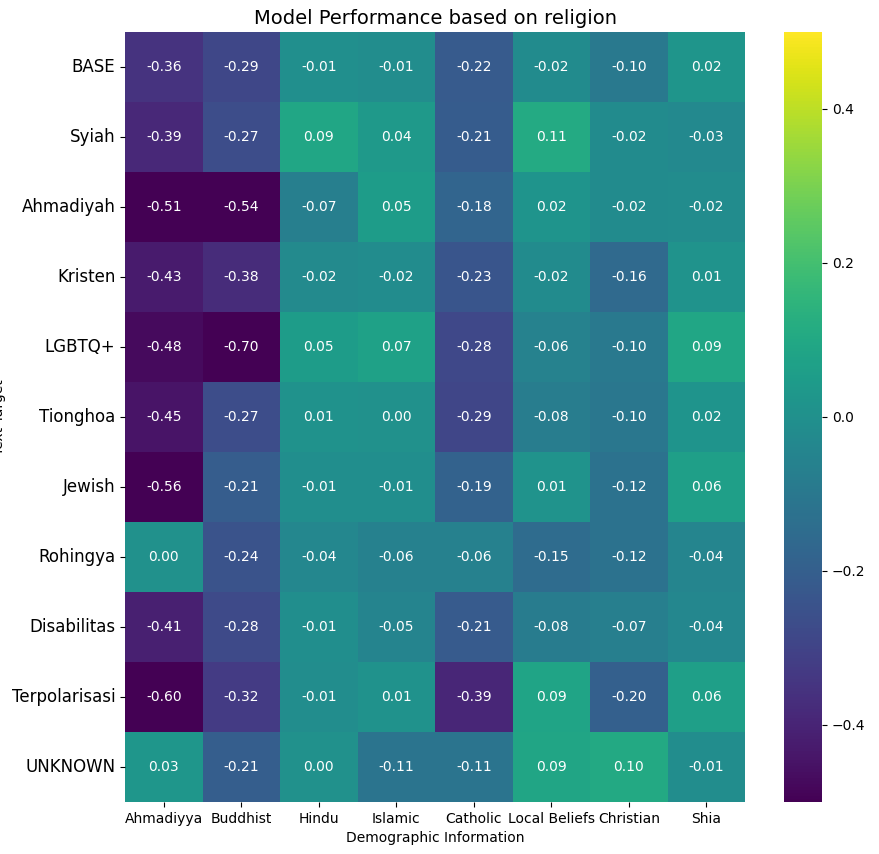}
    \vspace{-15pt}
    \caption{The effect ($\Delta$ F1) of giving religion information to IndoBERTweet on its hate speech text classification compared to not giving it any information.}  
    \label{fig:indo-demo-agama}
    \vspace{-15pt}
\end{figure}

\paragraph{However, demographic information hurts IndoBERTweet's classification performance.}
We attribute this to the diversity and subjectivity of hate speech texts. Though we hoped to increase the model's performance by providing demographic information, this instead fragmented the training data, with fewer training data for each demographic. In other words, we add another dimension(s) to the input without increasing the sample size. When the data is too few, this fragmentation hurts the model instead (Figure \ref{fig:indo-demo-agama}). Though the model learns the difference between each religion, there is very few data for some of them, such as \textbf{Ahmadiyya} and \textbf{Buddhist} which are the two least represented religion demographics in our dataset consisting of only 346 and 1,368 annotations respectively. Without this demographic information, the model only has the text as its input, which may explain the baseline's generally good performance.
\begin{figure}[h]
\vspace{-10pt}
    \centering
    \includegraphics[width=0.5\textwidth]{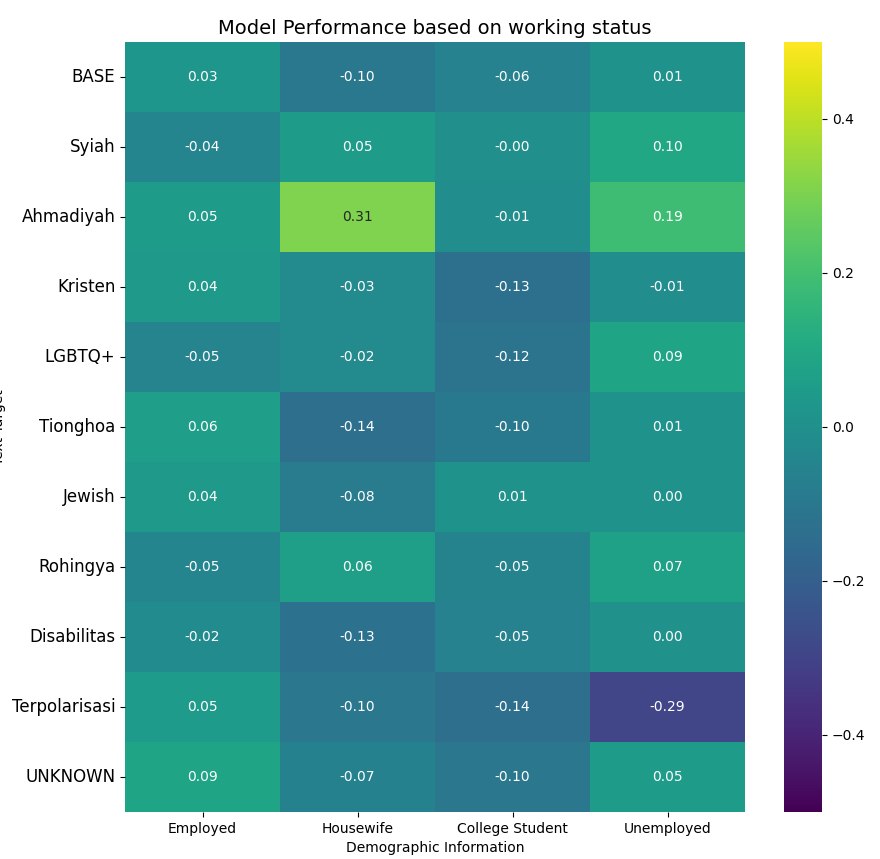}
    \vspace{-20pt}
    \caption{The effect ($\Delta$ F1) of giving topic information to IndoBERTweet on its hate speech text classification compared to not giving it any information.}  
    \label{fig:indo-topic-status kerja}
    \vspace{-15pt}
\end{figure}

\paragraph{Topic information increases IndoBERTweet's performance.} Focusing on the row entries from Figure \ref{fig:indo-topic-status kerja}, we observe that the model's performance increases in most cases. Though we hypothesized that the model would innately learn the topic of the text by itself, it failed to do so in practice, potentially due to a lack of data. Therefore, adding topic information improves the model's performance.
\section{Discussion}
The surge in online hate speech \cite{csis, aji-hatespeech-dashboard} poses societal risks \cite{hatespeech-williams-2019}. The challenge lies in defining hate speech and the debate over censoring such content. Some fear censorship could have a domino effect \cite{against1-censor-2013}, while others argue for immediate action due to existing societal damage \cite{for1-censor-2020}. Despite the complexity of defining hate speech, we can’t remain idle. We propose that researchers focus on creating hate speech detection systems focusing on protecting vulnerable minority groups, often targeted by hate speech and hate crimes, a crucial step for their protection.

To develop a hate speech detection system that filters texts targeting vulnerable minority groups, datasets with demographic information, like those in \citet{kumar-2021} and ours, are needed. However, such datasets are surprisingly scarce. Given the potential benefits to these groups, we call on researchers to promptly create similar datasets.

Furthermore, demographic datasets may have a larger role than expected. Our analysis indicates that large language models like gpt-3.5-turbo may contain biases. These biases can be reduced with metadata, which can form a “Persona” for effective model interaction. Our tests show that demographic information makes gpt-3.5-turbo a better hate speech detector. There could be other unexplored scenarios where demographic information is beneficial for different tasks.

\section*{Limitations}
Our work has the potential to pave the way for future research in creating better hate speech and toxic text detection. However, our work is not without flaws.

\paragraph{Baseline Performance Relies on a Private Dataset.} The baseline performance for the seven binary classification tasks was established by integrating our IndoToxic2024 dataset with a private dataset from \cite{csis}, accessed through collaboration.

\paragraph{Comparison of Fine-tuned and Zero-shot Models.} We compared the performance of a fine-tuned IndoBERTweet model against zero-shot gpt-3.5-turbo and SeaLLM-7B-v2.5 models. Due to the extensive number of experiments conducted with gpt-3.5-turbo, we were unable to fine-tune it for a direct comparison. Additionally, we discontinued the use of SeaLLM-7B-v2.5 due to its subpar performance relative to the other models.

\paragraph{Inconsistent Annotator Count Across Annotation Phases.} The varying number of annotators across different annotation phases may lead to inconsistencies in the data distribution. This could potentially result in an incomplete representation of the overall demographic traits, as not all demographic dimensions may be uniformly captured.

\paragraph{Controlled Topic Distribution in Main Annotation Phase 1.} Annotators did not received a completely random texts when they annotate the 1000-text batch in main annotation phase 1. Instead, they received 50\% text \textit{that} mentions their group, and 50\% random text that \textit{did not} mentions their groups. This applies to all dataset except the Shia \& Ahmadiyya dataset,who received only 39.3\% text of their group due to the scarcity of existing data.

\paragraph{Use of a Naive Keyword-based Approach for Topic Extraction.} Our approach to extracting featured topics relies on predefined keywords, which may overlook nuanced or emergent themes. This limitation could restrict the scope of our analysis and prevent the identification of more subtle or complex topics present in the text. A more dynamic and flexible topic extraction approach could potentially enhance the richness and accuracy of topic identification within IndoToxic2024.

\section*{Ethics Statement}
The creation of this dataset exposes annotators to potentially harmful hate speech texts. To avoid excessive mental strain, we intentionally extended the annotation duration to two and a half months. Individuals are preemptively warned and asked for consent during the initial recruitment process. Furthermore, annotators are permitted to quit the annotation of texts if they feel unable to proceed. We recognize the potential misuse of such datasets, which could include training models to generate more hate speech. Yet, it’s worth noting that without these datasets, it is alarmingly straightforward to train a model to produce toxic content, as the source of their training data, the internet, still consists of hate speech and toxic texts. This has been demonstrated by numerous researchers who have attempted to reduce toxic output or identify vulnerabilities in large language models (refer to \citet{easy-toxic-1, easy-toxic-2}). On the other hand, the area of developing models to detect hate speech targeted at specific demographic groups is still green, with a notable lack of available data, especially in Indonesia. Weighing these considerations, we firmly believe that the potential benefits of this type of dataset significantly outweigh the possible misuse.

\section*{Acknowledgements}
We thank the Boston University Computing Facility for lending us their computing unit. 
This work is partly supported by the Alliance of Independent Journalists, the Monash Incubator Grant, and Monash Data Future Institute Research Grants. Any opinions, findings, conclusions, or recommendations expressed here are those of the authors and do not necessarily reflect the view of the sponsor.

\bibliography{anthology,references}
\bibliographystyle{acl_natbib}

\onecolumn
\newpage
\section*{Appendix} 
\appendix
\section{Example of Divisive Texts}
\label{app:divisive-text}

A list of divisive text is available in Table \ref{tab:divisive-text}.

\begin{table*}[ht]
\centering
\small
\begin{tabular}{p{8.5cm}|p{6.5cm}}
\hline
\textbf{Divisive Text} & \textbf{Comment} \\
\hline
Around 800 to 1000 Christians still live in Gaza, which is one of the oldest Christian communities worldwide. They are also evicted and killed. Entire faith groups died out in the land where Christianity originated. & \textbf{3 of 3 Christians or Catholics}, and \textbf{0 of 10 non-Christians or Catholics} annotators annotate this text as hate speech. \\
\hline
Be a woman who is crazy about work. Because there are enough women crazy about boys. Even though people keep seeing us hanging out. & \textbf{1 of 4 undergraduates} and \textbf{3 of 3 high school graduates }annotators annotate this text as hate speech. However, \textbf{2 of 5 females} and \textbf{2 of 3 males} annotators annotate this text as hate speech. \\
\hline
"GANJA(Weed) FOR ALL CEBONG" with all their 45 spirit putting up billboards, they forgot one character and turned it into something completely different... hahahaah, this is what happens when you want to scam people. & \textbf{2 of 3 Ganjar voters (Group 3)}, \textbf{5 of 5 Prabowo voters (Group 2)}, and \textbf{0 of 3 Anies voters (Group 1)} annotate this text as hate speech.\\
\hline
Everybody knows that most PDIP supporters are Christians and that anyone gaining the support of PDIP will also get the support of Christians. However, this chances because of Ganjar Pranowo's stance on banning Israel's soccer team from coming to Indonesia to compete, which makes Christians unhappy. & \textbf{3 of 4 males} and \textbf{0 of 7 females} annotators annotate the text as hate speech. \\
\hline
I used to praise Jokowi but not anymore, why? Because he used his child for evil people. If Gibran was made to become a candidate for president, it would have been okay. But, Gibran was made as a candidate for vice president, used only to gain supporters for Golkar political group. & \textbf{7 of 8 non-Chinese} and \textbf{0 of 3 Chinese} annotators annotate this text as hate speech. \\
\hline
Goooo GaMa (Word play on the president and vice president candidate of Group 3)! Legally defective products will continue to create defective products. & \textbf{6 of 7 Gen Z} and \textbf{0 of 4 Gen X and Millenials} annotate this text as hate speech.\\
\end{tabular}
\caption{Examples of divisive texts and the demographic group in which they are divisive.}
\label{tab:divisive-text}
\vspace{-6pt}
\end{table*}
\section{Keywords Used for Scraping}
\label{sec:scrape-keywords}
cina, china, tionghoa, chinese, cokin, cindo, chindo, shia, syiah, syia, ahmadiyya, ahmadiyah, ahmadiya, ahmadiyyah, transgender, queer, bisexual, bisex, gay, lesbian, lesbong, gangguan jiwa, gangguan mental, lgbt, eljibiti, lgbtq+, lghdtv+, katolik, khatolik, kristen, kris10, kr1st3n, buta, tuli, bisu, budek, conge, idiot, autis, orang gila, orgil, gila, gendut, cacat, odgj, zionis, israel, jewish, jew, yahudi, joo, anti-christ, anti kristus, anti christ, netanyahu, setanyahu, bangsa pengecut, is ra hell, rohingya, pengungsi, imigran, sakit jiwa, tuna netra, tuna rungu, sinting.
\section{Annotation Guidelines}
\label{sec:annotguide}

\subsection{Definition}
\textbf{Toxic comments}\quad is a post, text, or comment that is harsh, impolite, or nonsensical, causing you to become silent and unresponsive, or that is filled with hatred and aggression, provoking feelings of disgust, anger, sadness, or humiliation, making you want to leave the discussion or give up sharing your opinion.

\textbf{Profanity or Obscenity}\quad The message / sentence on social media posts contains offensive, indecent, or inappropriate in a way that goes against accepted social norms. It often involves explicit or vulgar language, graphic content, or inappropriate references. Essentially, it's a message that is likely to be considered offensive or objectionable by most people.

\textbf{Threat / Incitement to Violence}\quad The message / sentence on social media posts conveys an intent to cause harm, danger, or significant distress to an individual or a group. It often includes explicit or implicit threats of violence, physical harm, intimidation, or any action that creates a sense of fear or apprehension.

\textbf{Insults}\quad The message / sentence on social media posts contains offensive, disrespectful, or scornful language with the intention of belittling, offending, or hurting the feelings.

\textbf{Identity Attack}\quad The message / sentence on social media posts deliberately targets and undermines a person's sense of self, identity, or personal characteristics. This can include derogatory comments, or harmful statements aimed at aspects such as one's race, gender, sexual orientation, religion, appearance, or other defining attributes.

\textbf{Sexually Explicit}\quad The message / sentence on social media posts contains explicit and detailed descriptions or discussions of sexual activities, body parts, or other related content.

\subsection{Manual Annotation}
\textbf{Q1: Does this text appear to be random spam or lack context? }
\begin{compactitem}
    \item Yes
    \item No
\end{compactitem}
\textbf{Q2: Does this text related to Indonesian 2024 General Election? }
\begin{compactitem}
    \item Yes
    \item No
\end{compactitem}
\textbf{Q3: Does this text contain toxicity (hate speech)? }\\
\textit{Note}: Irrelevant toxicity or hate speech includes hate speech that is meant as a joke among friends or is not considered hate speech by the recipient. Thus, it will be coded as "No".
\begin{compactitem}
    \item Yes
    \item No
\end{compactitem}

\textbf{Q4: What is the type of toxicity?}\\
\textit{Note:} Code up to two or more types. Consider the following sentences as an example: \textit{“PDIP Provokasi Massa pendukungnya geruduk kediaman Anies”}. This headline should be coded as both  threat and incitement to violence.\\
\textbf{Q4-1: Does the message contains profanity/obscenity?}
\begin{compactitem}
    \item Yes
    \item No
\end{compactitem}
\textbf{Q4-2: Does the message contain threat / incitement to violence?}
\begin{compactitem}
    \item Yes
    \item No
\end{compactitem}
\textbf{Q4-3: Does the message contain insults?}
\begin{compactitem}
    \item Yes
    \item No
\end{compactitem}
\textbf{Q4-4: Does the message contain an identity attack?}
\begin{compactitem}
    \item Yes
    \item No
\end{compactitem}
\textbf{Q4-5: Does the message contain sexually explicit?}
\begin{compactitem}
    \item Yes
    \item No
\end{compactitem}



\section{Mapping of \textbf{Keywords-to-Topics}}
\label{sec:mapping_topics}

\begin{compactitem}
\item \textbf{Shia} : shia, syia, syiah
\item \textbf{Ahmadiyya} : ahmadiya, ahmadiyah, ahmadiyya, ahmadiyyah
\item \textbf{Christian} : anti christ, anti kristus, anti-christ, kris10, kristen, kr1st3n, katolik, khatolik
\item \textbf{LGBTQ}+ : bisex, bisexual, eljibiti, gay, lesbian, lesbong, lgbt, lgbtq+, lghdtv+, queer, transgender
\item \textbf{Chinese} : china, chindo, chinese, cina, cindo, cokin, tionghoa
Jewish : is ra hell, israel, jew, jewish, joo, netanyahu, netanhayu, setanyahu, yahudi, zionis, bangsa pengecut
\item \textbf{Rohingya} : rohingya, imigran, pengungsi
\item \textbf{Disability} : odgj, idiot, autis, bisu, budek, buta, cacat, gangguan jiwa, gangguan mental, gila, ogdj, sinting, orang gila, orgil, sakit jiwa, tuli, tuna netra, tuna rungu, conge, gendut
\end{compactitem}

\newpage
\section{Prompt for Synthetic Text Generation using gpt-3.5-turbo}
\label{app:synthetic-gen-prompt}
\begin{figure}[h]
    \centering
    \includegraphics[width=0.5\textwidth]{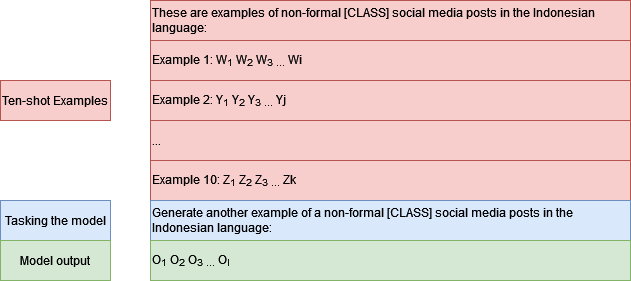}
    \caption{The template we use to prompt gpt-3.5-turbo through ten-shot prompting.}  
    \label{fig:synth-gen}
    \vspace{-15pt}
\end{figure}

To generate synthetic data from gpt-3.5-turbo to help increase model performance for various binary classification task, we utilize the prompt visualized in Figure \ref{fig:synth-gen}

\section{Within \& Between Group ICR Score Calculation}
\label{sec:rigorICR}

To compute the toxicity ICR score for a demographic group, we calculate the weighted average of Gwet’s AC1 score for every pairwise combination of annotators within the same group, using the volume of text in each pair as the weight. This approach maximizes the utilization of available data. A similar calculation method is implemented to find the ICR score between two groups, with the modification that each pair consists of members from different groups. We refer to these metrics as “within-group ICR score” and “between-group ICR score” respectively.

We can rigorously define an equation to compute ICR score within a group as
\begin{equation*}
    \small
     \gamma(g)= \frac{\sum\limits_{i,j\in g}dim(v_{ij})\cdot\text{Gwet}(\phi_i(v_{ij}),\phi_j(v_{ij}))}{\sum\limits_{i,j\in g}dim(v_{ij})}
\end{equation*}

Where $g$ are an arbitrary groups in a demographic; $v_{ij}$ are set of text that both mutually by annotators $i$,$j$; and $\phi_i,\phi_j$ are annotation result from $i,j$. To calculate ICR score between two groups, we slightly modified the equation above into

\begin{equation*}
    \small
     \Gamma(g_1,g_2)= \frac{\sum\limits_{i\in g_1,j \in g_2}dim(v_{ij})\cdot\text{Gwet}(\phi_i(v_{ij}),\phi_j(v_{ij}))}{\sum\limits_{i\in g_1,j \in g_2}dim(v_{ij})}
\end{equation*}

Where $g_1$,$g_2$ are arbitrary two groups in a demographic; $v_{ij}$ are set of text that both mutually by annotators $i$,$j$; and $\phi_i,\phi_j$ are annotation result from $i,j$.
\newpage
\section{Performance of Fine-tuned Models on Multiple Topics}
\label{app:f1_demotopic}

The figures below show the performance of our fine-tuned models, trained per demographic group, on various text topics. As the demographic group changes, the performance of each model also differs per topic.

\begin{figure}[h]
    \centering
    \includegraphics[width=1\textwidth]{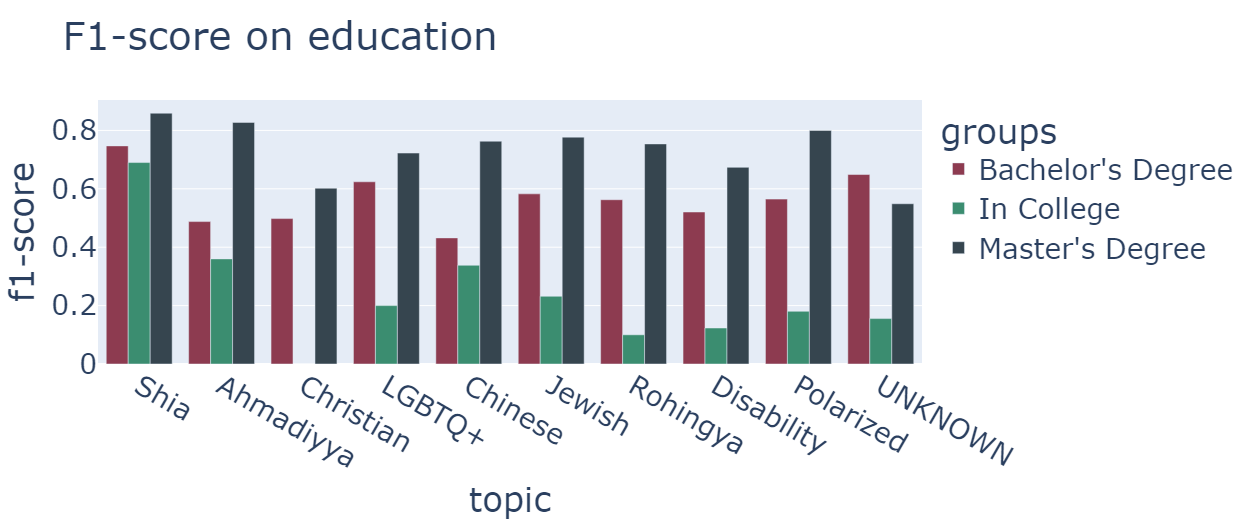}
    \label{fig:demotopic-education}
\end{figure}
\begin{figure}[h]
    \centering
    \includegraphics[width=1\textwidth]{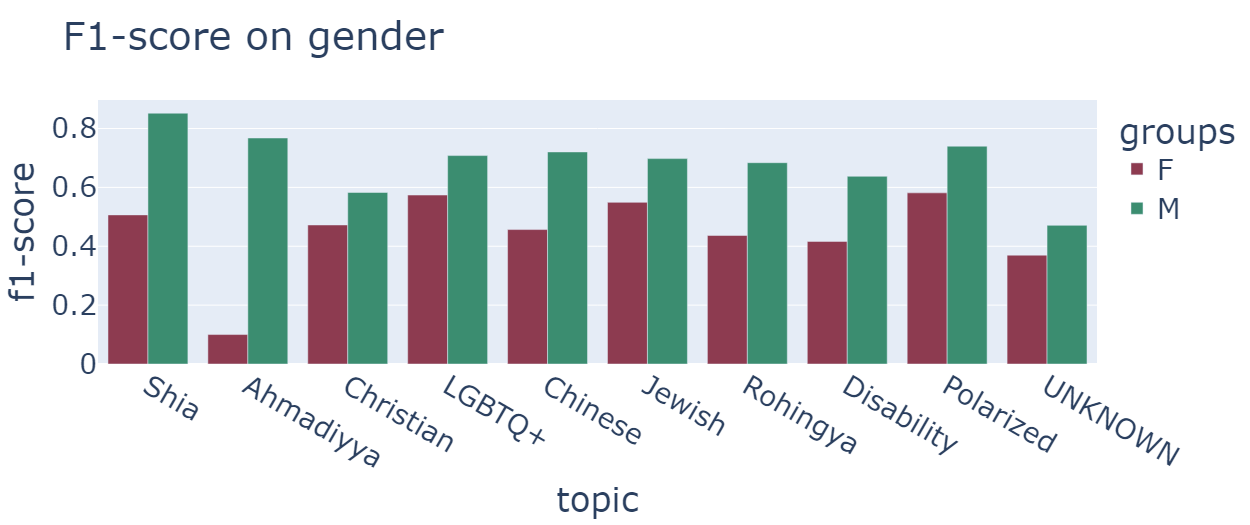}
    \label{fig:demotopic-gender}
\end{figure}
\begin{figure}[h]
    \centering
    \includegraphics[width=1\textwidth]{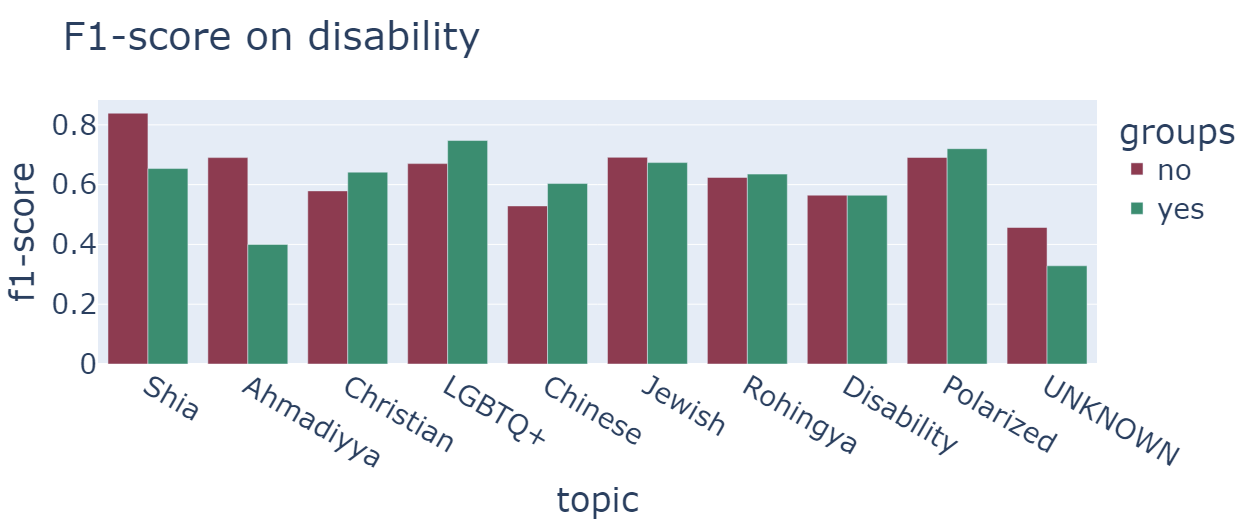}
    \label{fig:demotopic-disability}
\end{figure}
\begin{figure}[h]
    \centering
    \includegraphics[width=1\textwidth]{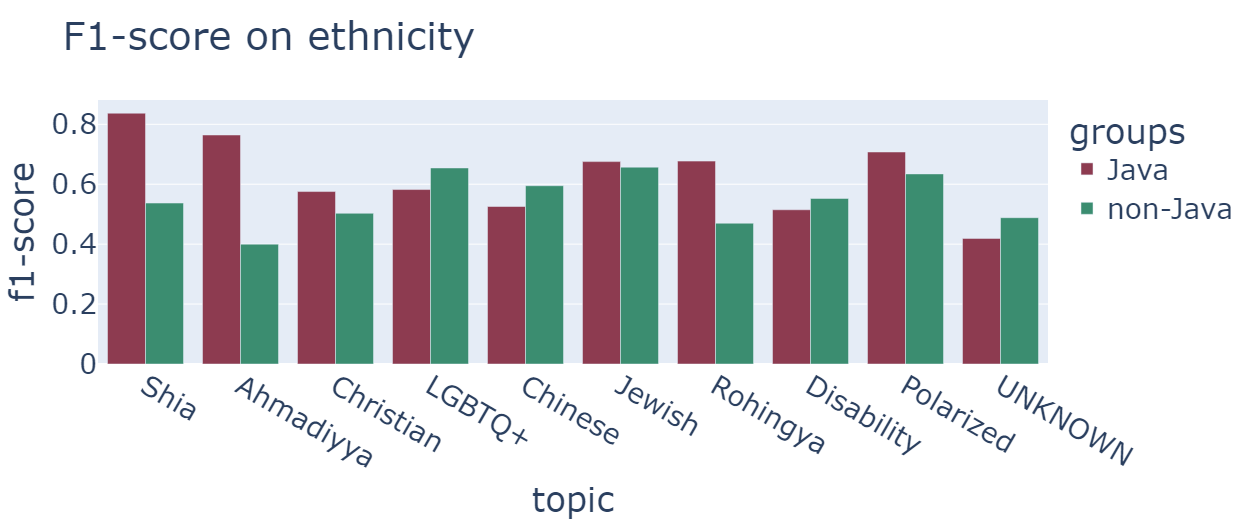}
    \label{fig:demotopic-ethnicity}
\end{figure}

\newpage
\newpage
\newpage
\section{Effect of Demographic and Topic Information to Model Performance}
\label{app:heatmaps}

\subsection{gpt-3.5-turbo}
\begin{figure}
    \centering
    \includegraphics[width=1\textwidth]{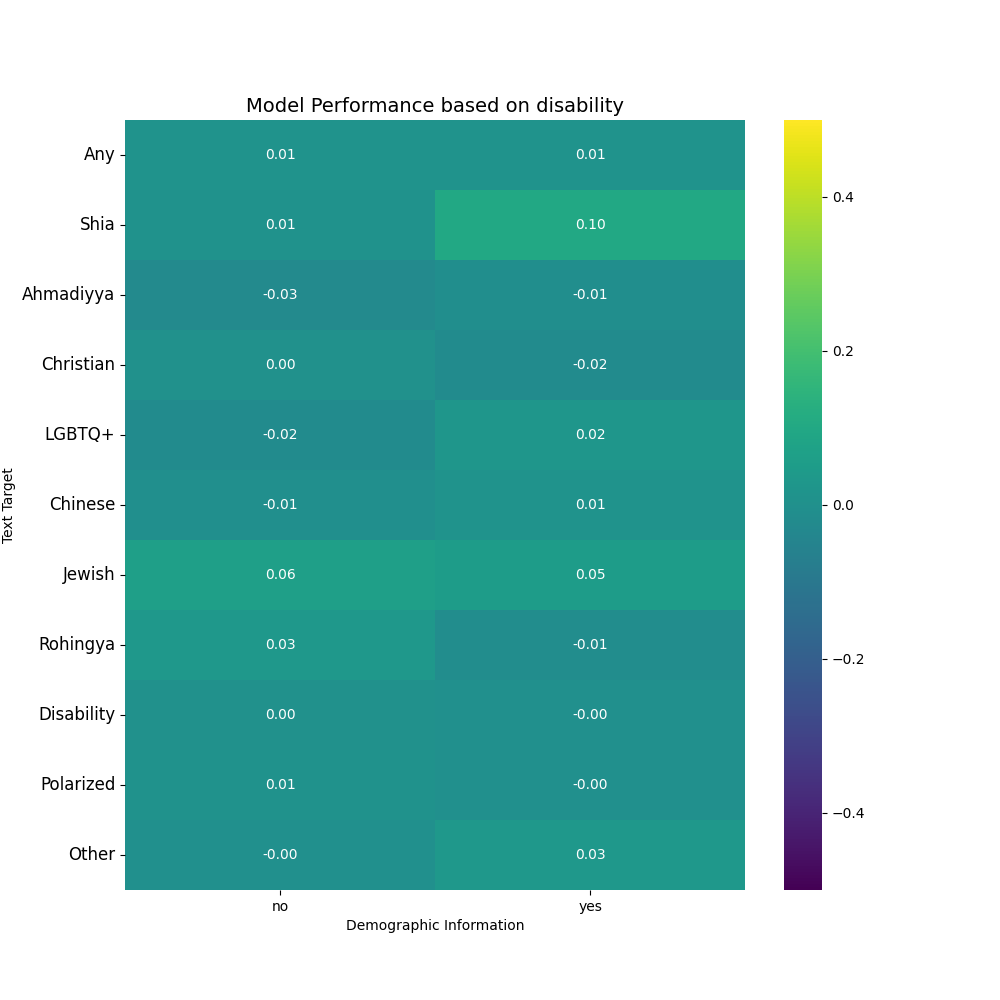}
\end{figure}
\begin{figure}
    \centering
    \includegraphics[width=1\textwidth]{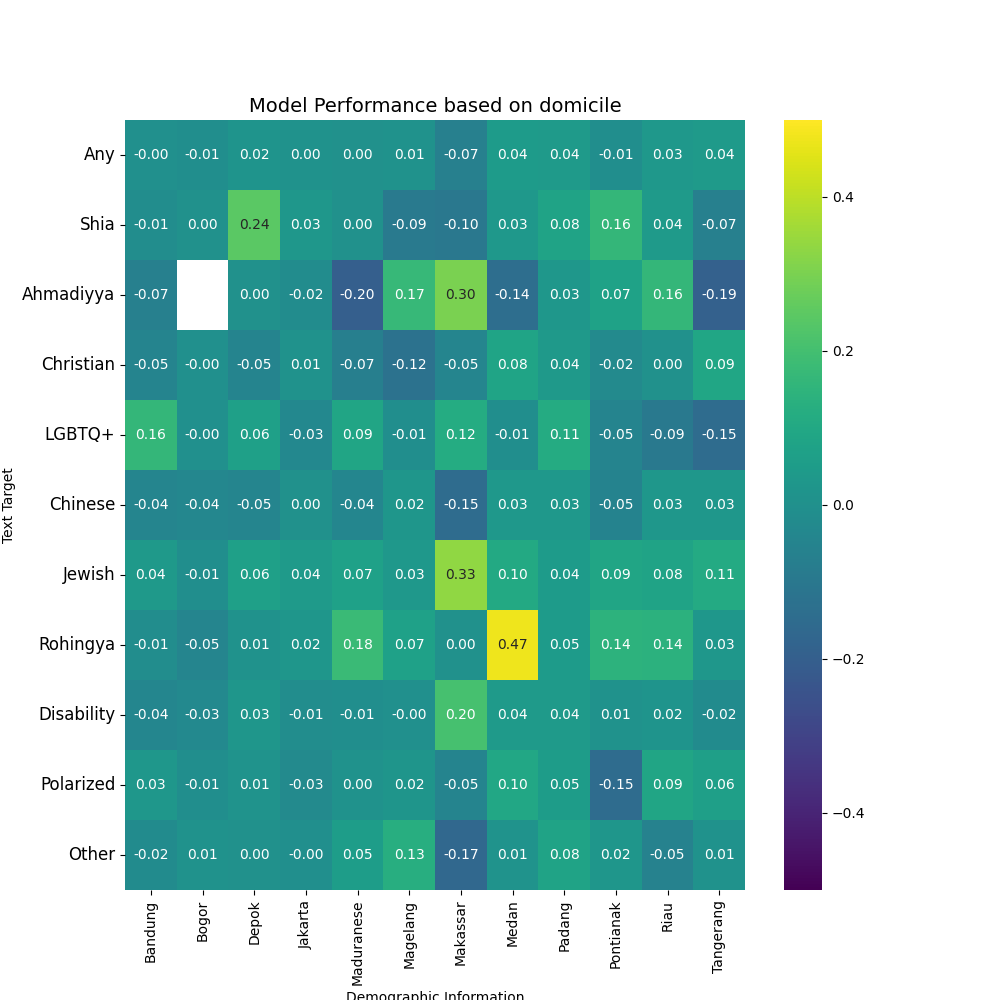}
\end{figure}
\begin{figure}
    \centering
    \includegraphics[width=1\textwidth]{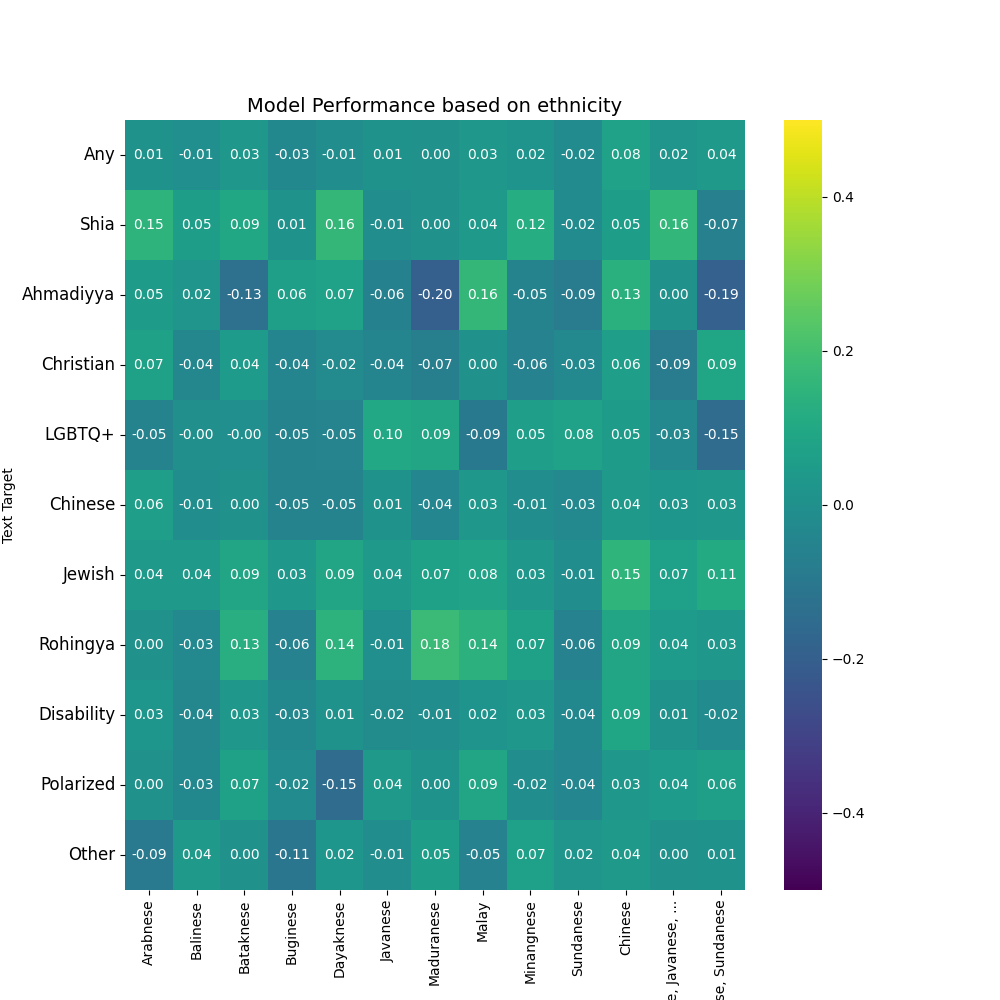}
\end{figure}
\begin{figure}
    \centering
    \includegraphics[width=1\textwidth]{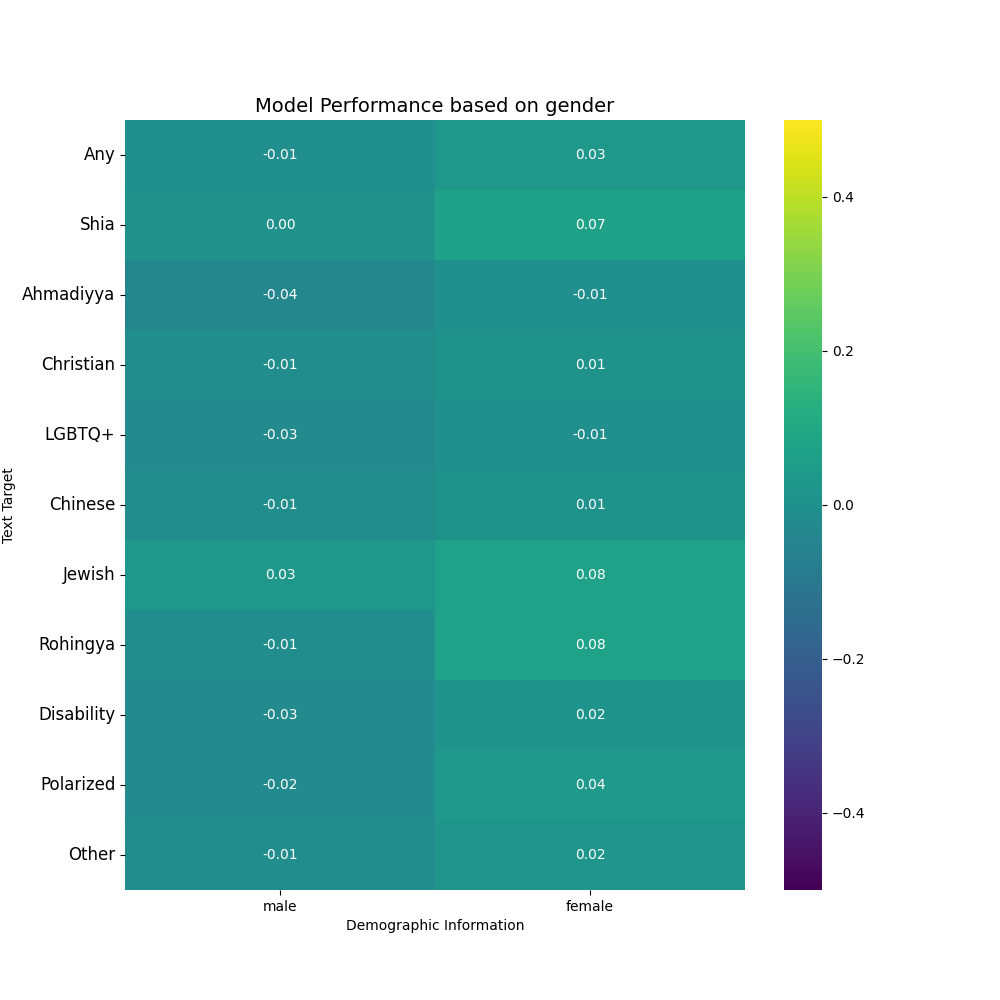}
\end{figure}
\begin{figure}
    \centering
    \includegraphics[width=1\textwidth]{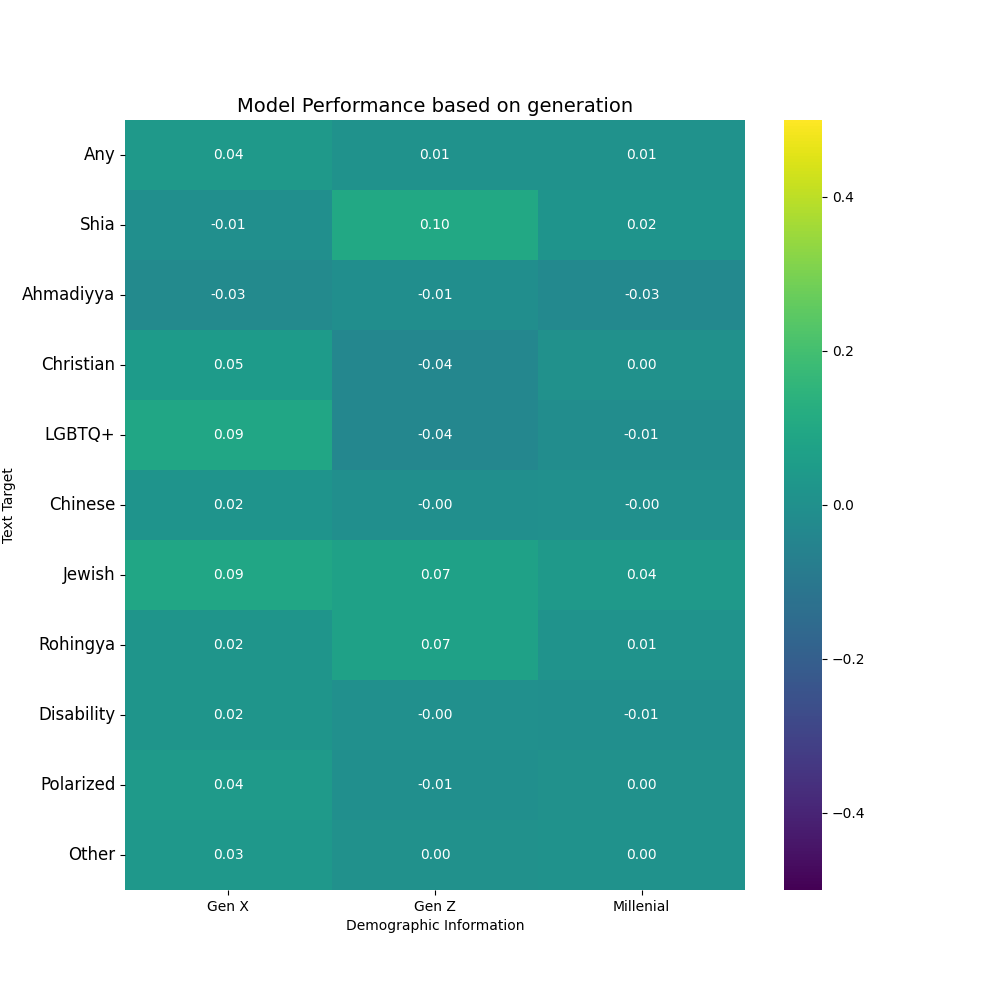}
\end{figure}
\begin{figure}
    \centering
    \includegraphics[width=1\textwidth]{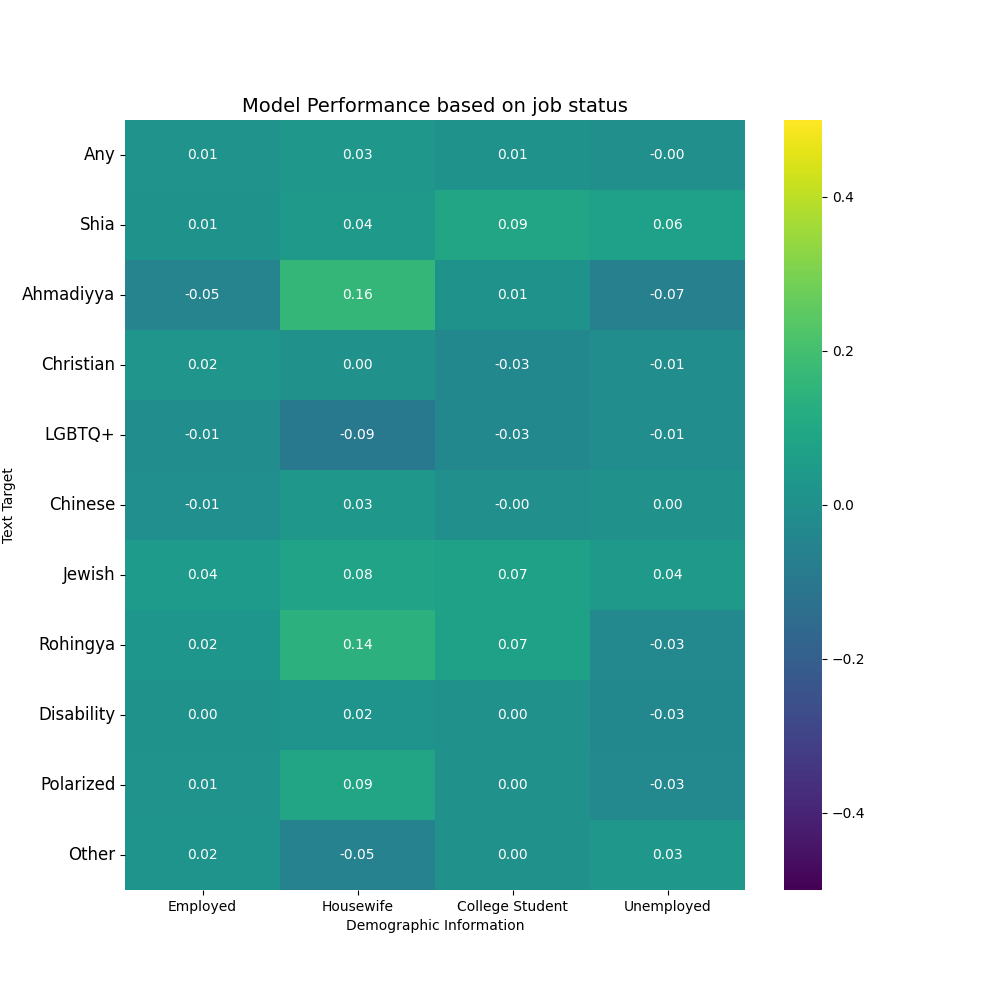}
\end{figure}
\begin{figure}
    \centering
    \includegraphics[width=1\textwidth]{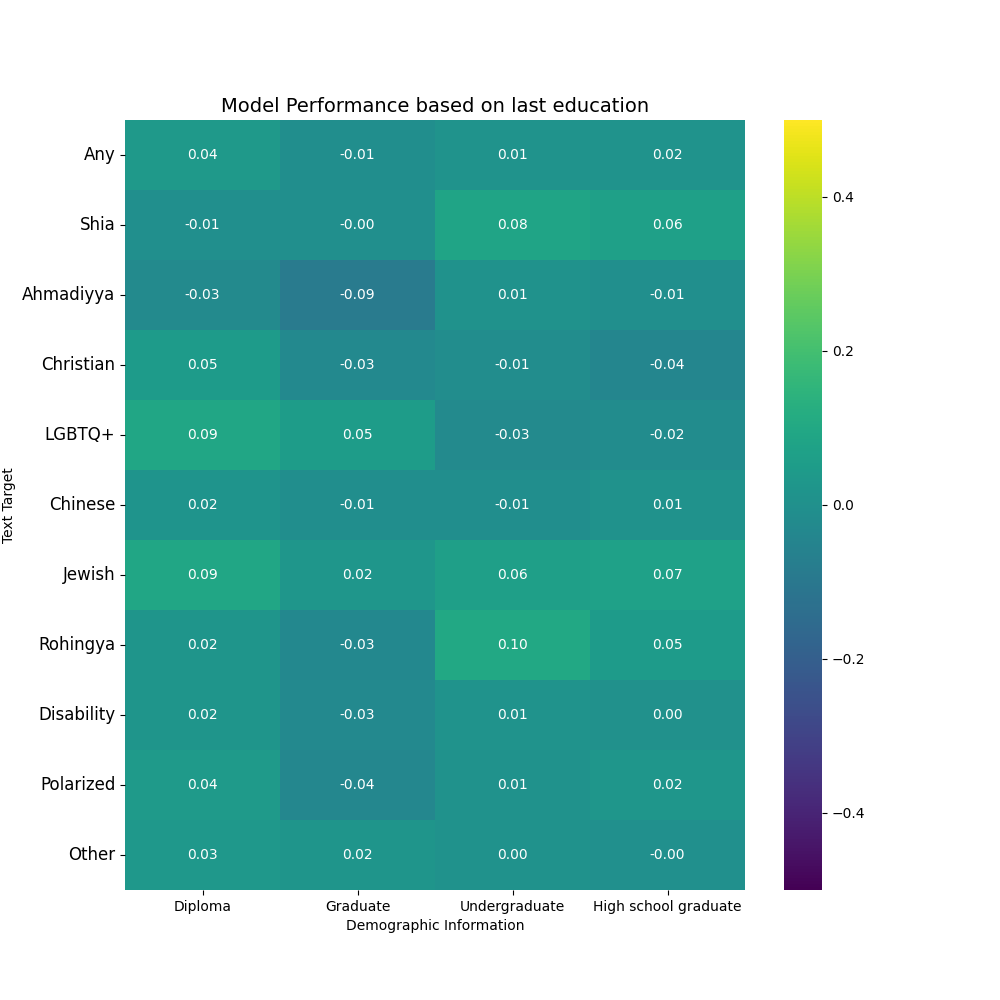}
\end{figure}
\begin{figure}
    \centering
    \includegraphics[width=1\textwidth]{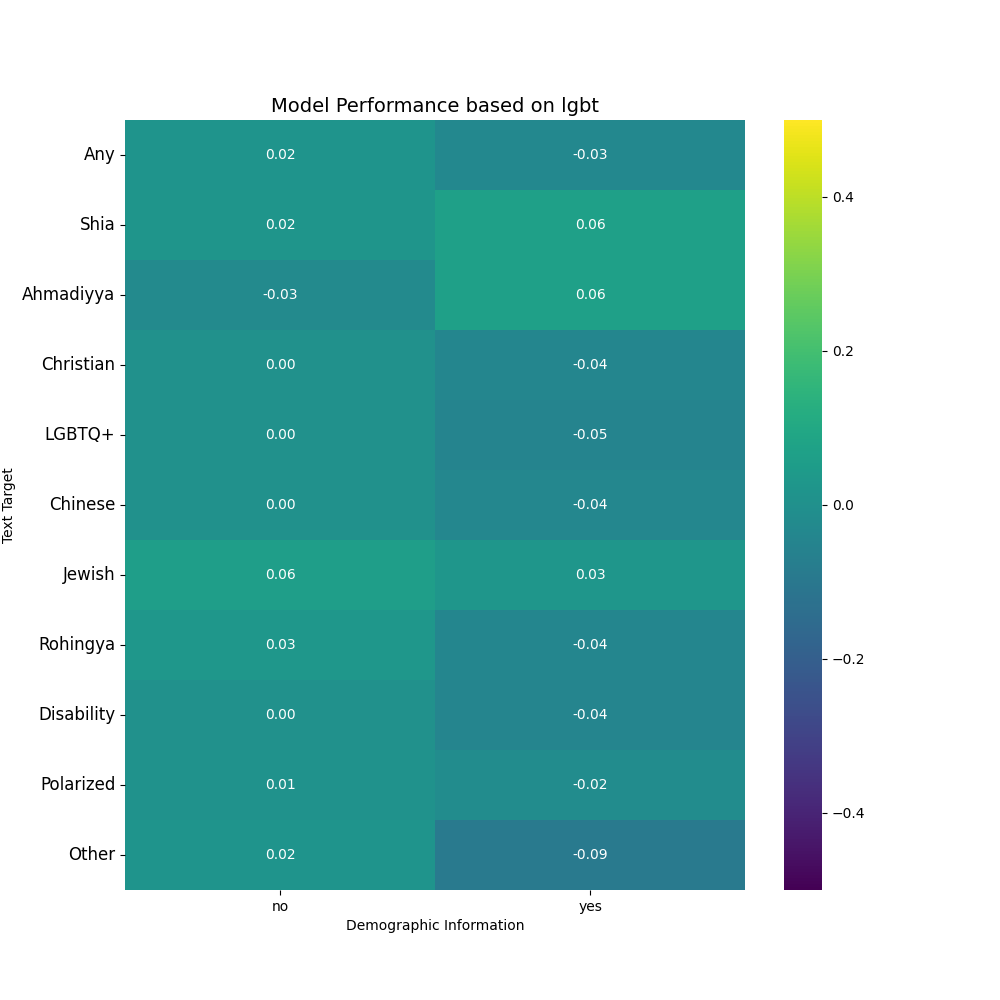}
\end{figure}
\begin{figure}
    \centering
    \includegraphics[width=1\textwidth]{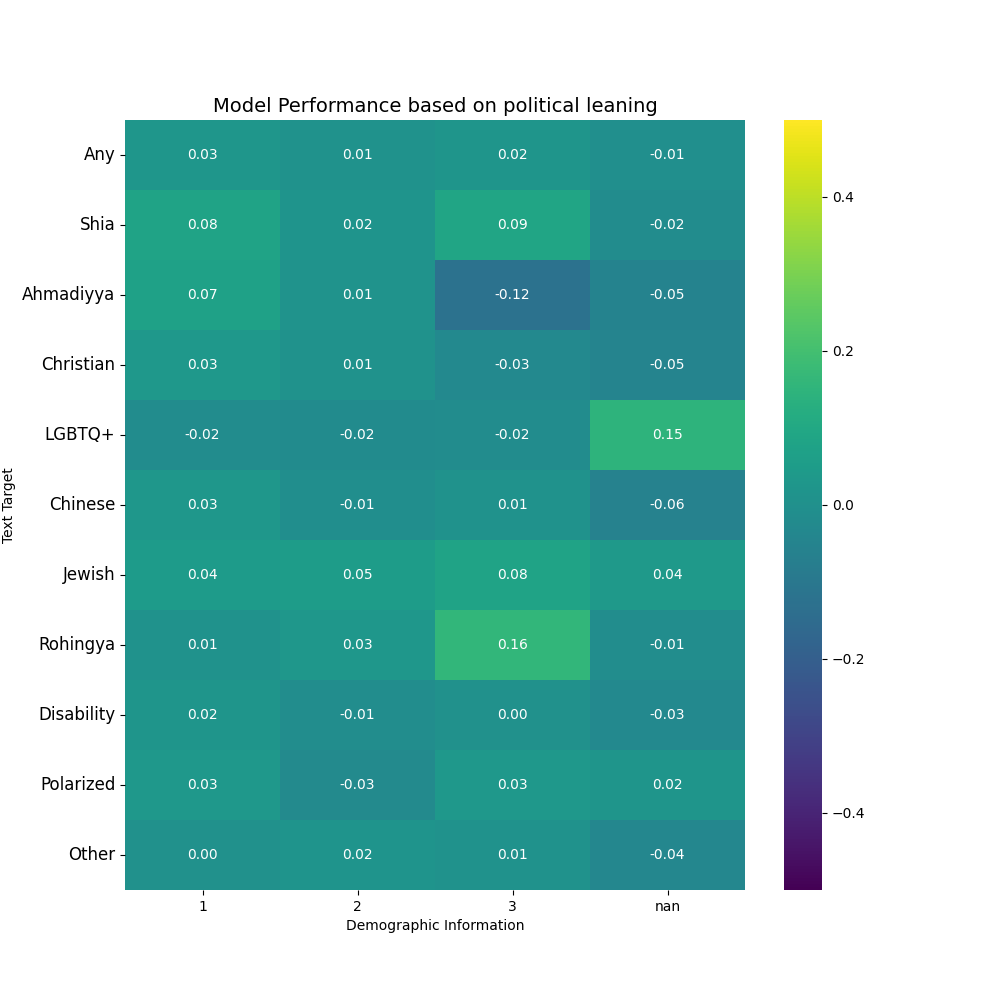}
\end{figure}
\begin{figure}
    \centering
    \includegraphics[width=1\textwidth]{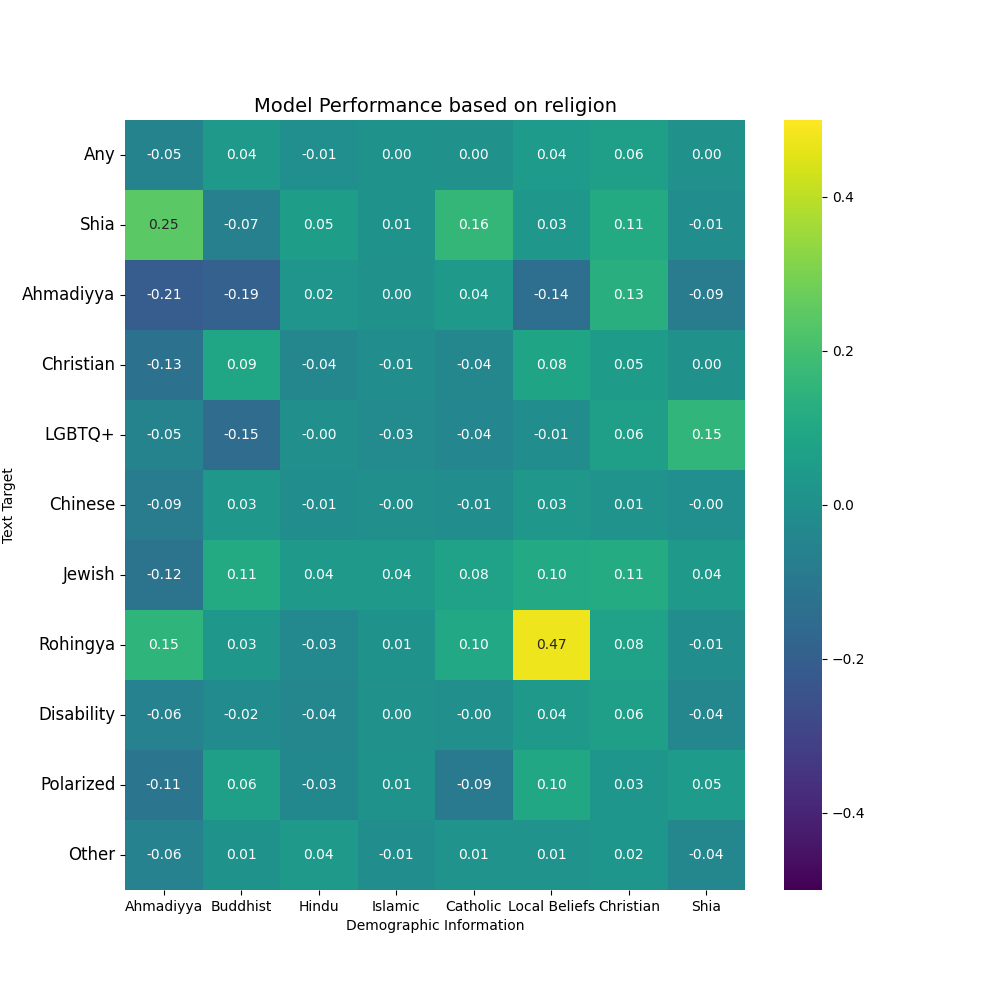}
\end{figure}

\newpage
\subsection{IndoBERTweet with Demographic Information}
\begin{figure}[H]
    \centering
    \includegraphics[width=1\textwidth]{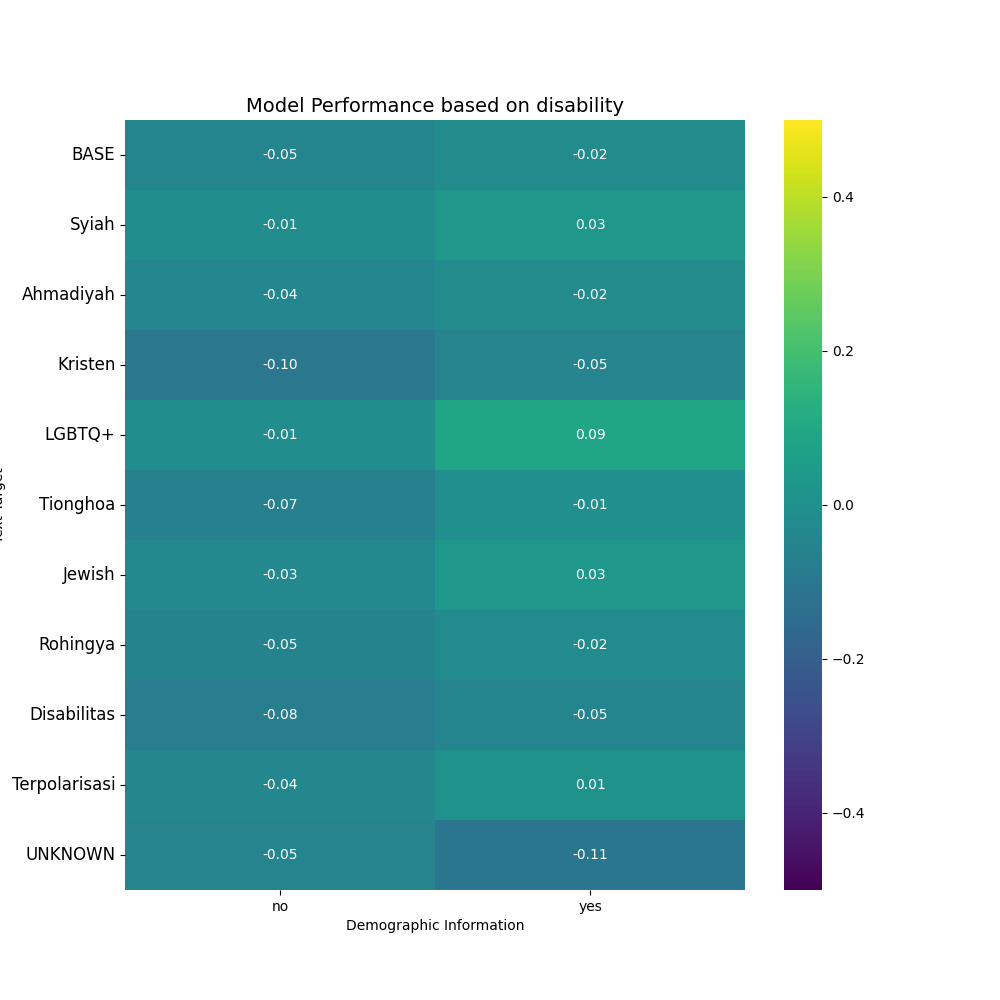}
\end{figure}
\begin{figure}
    \centering
    \includegraphics[width=1\textwidth]{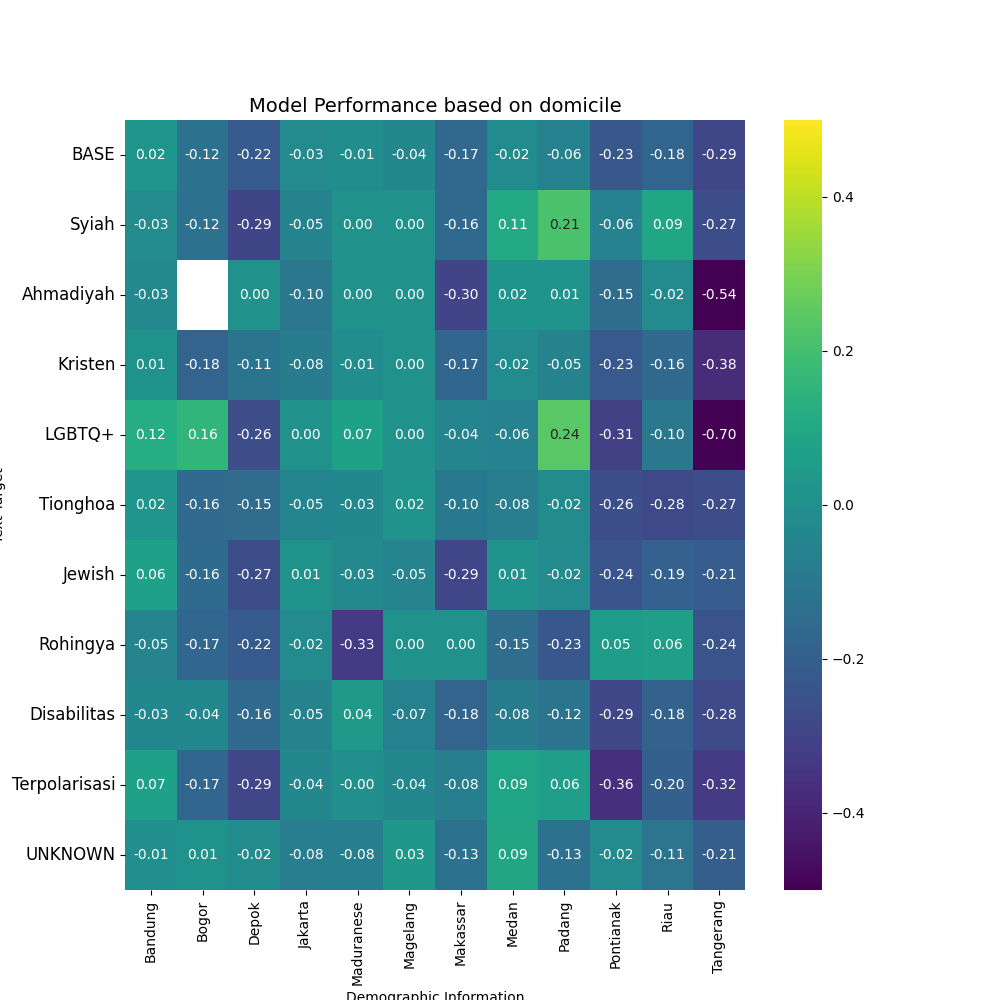}
\end{figure}
\begin{figure}
    \centering
    \includegraphics[width=1\textwidth]{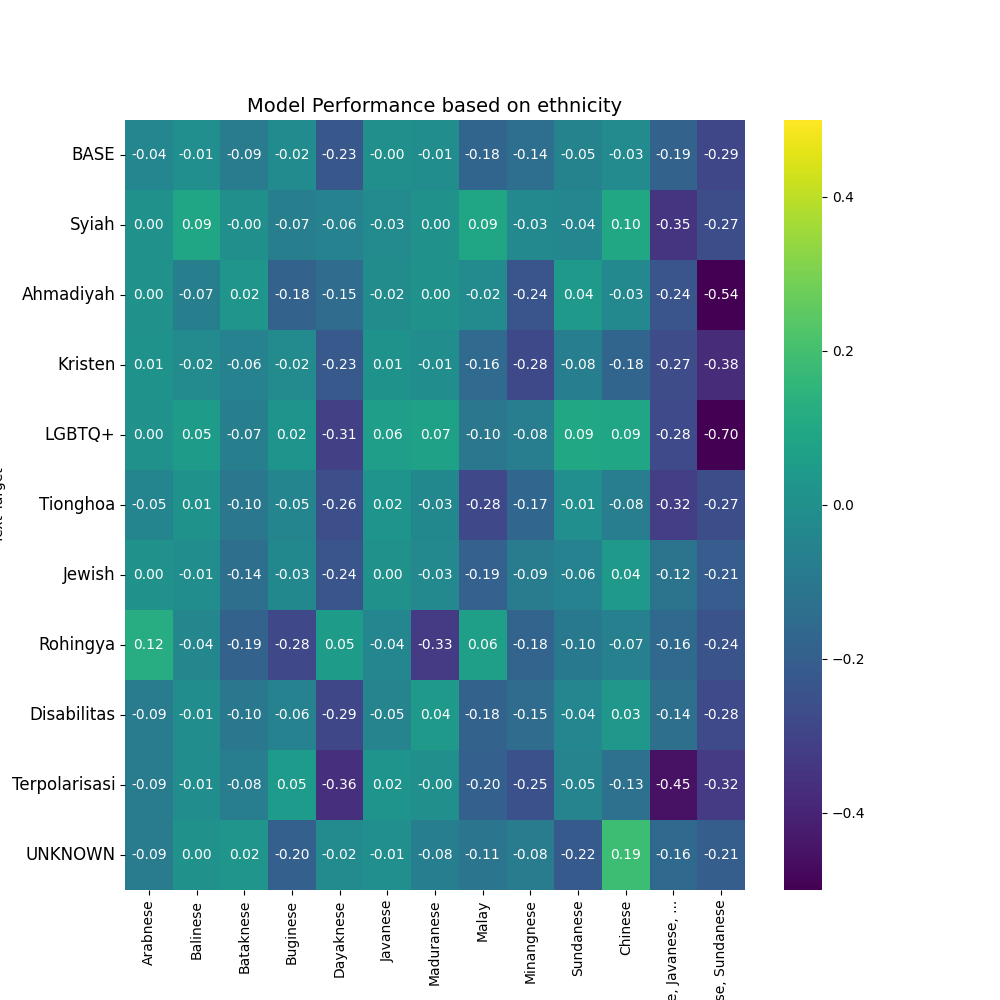}
\end{figure}
\begin{figure}
    \centering
    \includegraphics[width=1\textwidth]{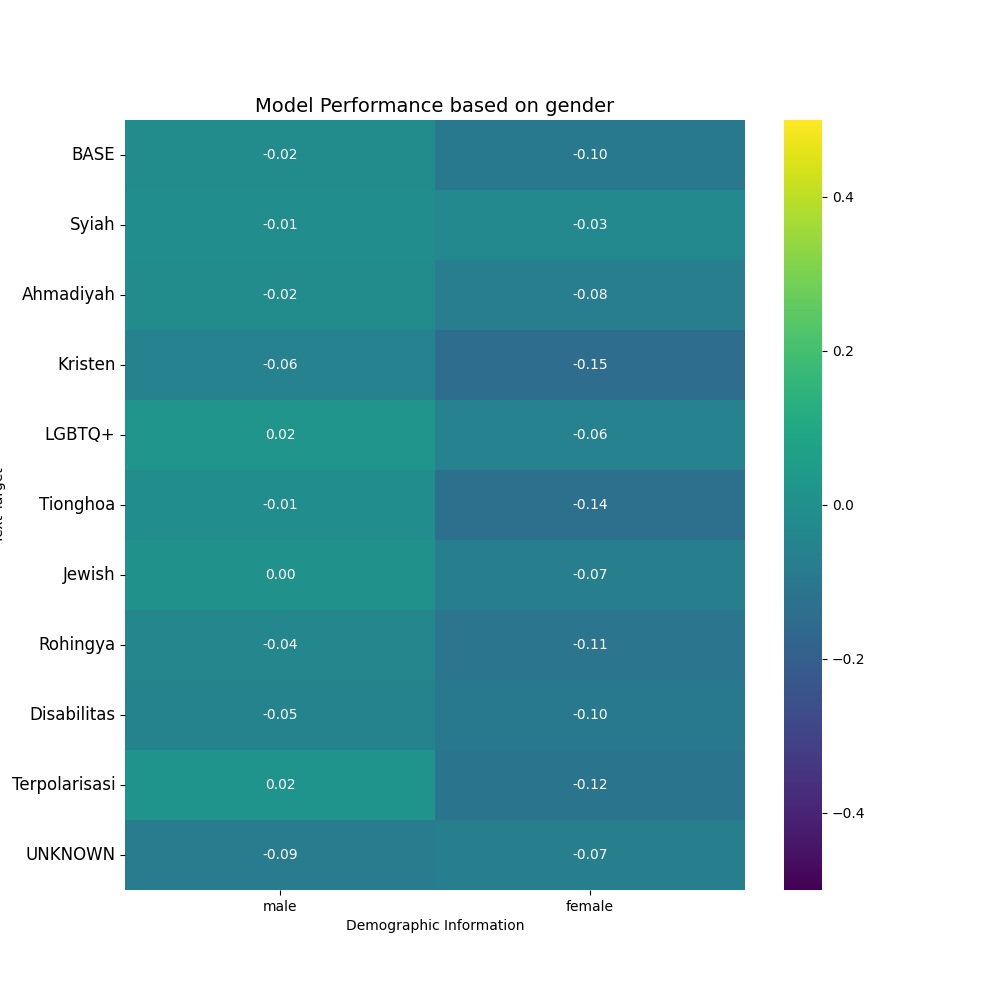}
\end{figure}
\begin{figure}
    \centering
    \includegraphics[width=1\textwidth]{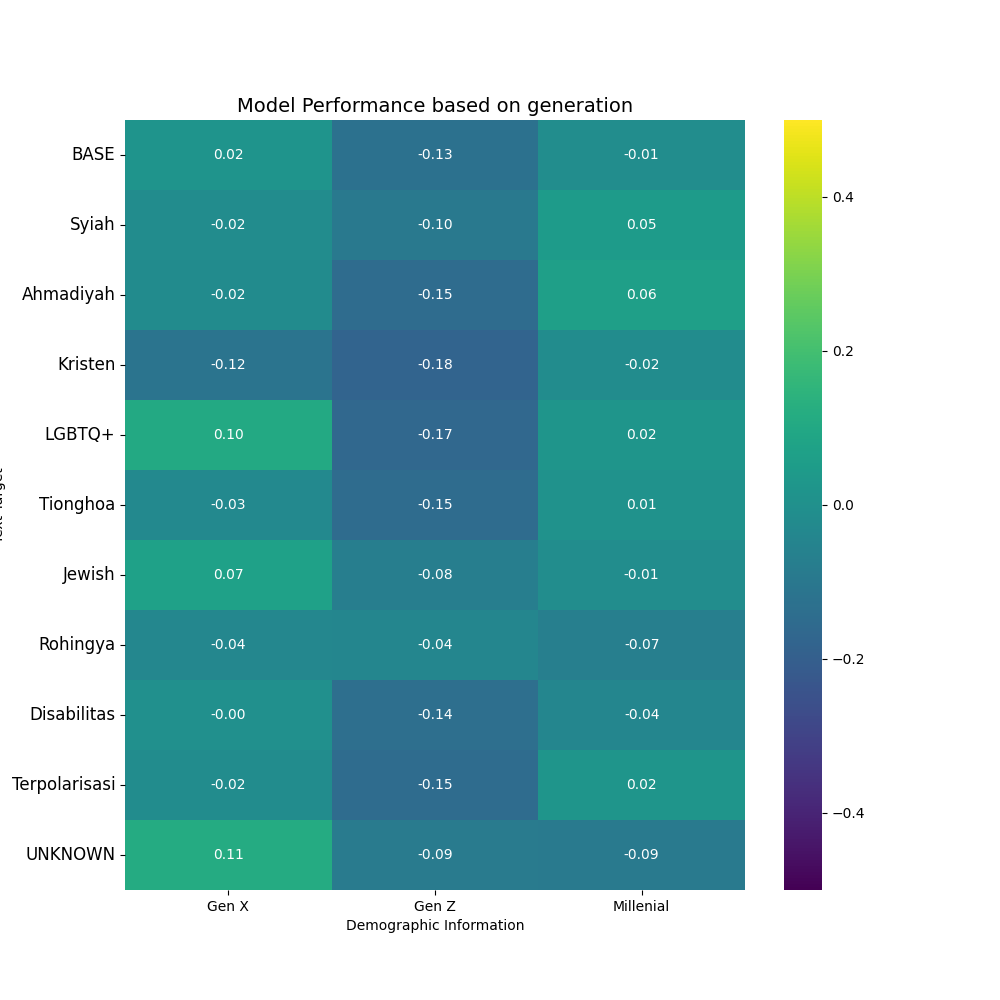}
\end{figure}
\begin{figure}
    \centering
    \includegraphics[width=1\textwidth]{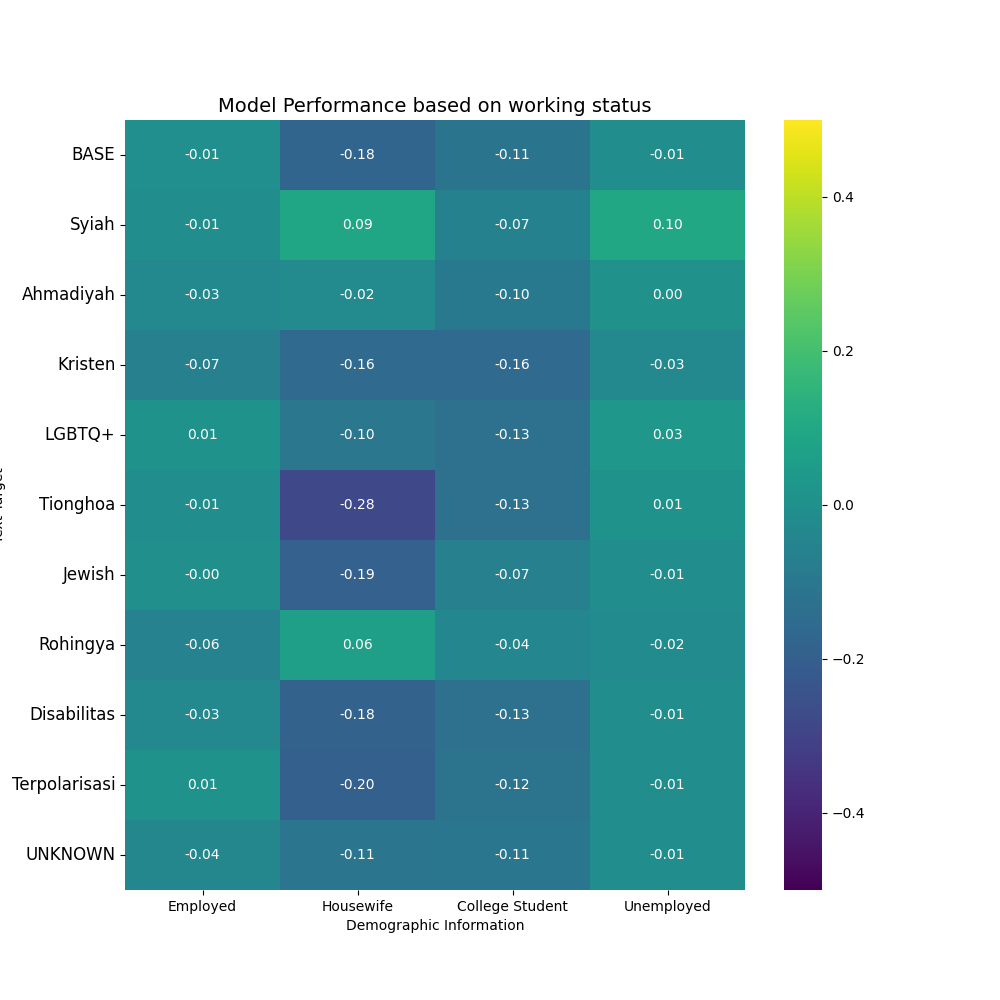}
\end{figure}
\begin{figure}
    \centering
    \includegraphics[width=1\textwidth]{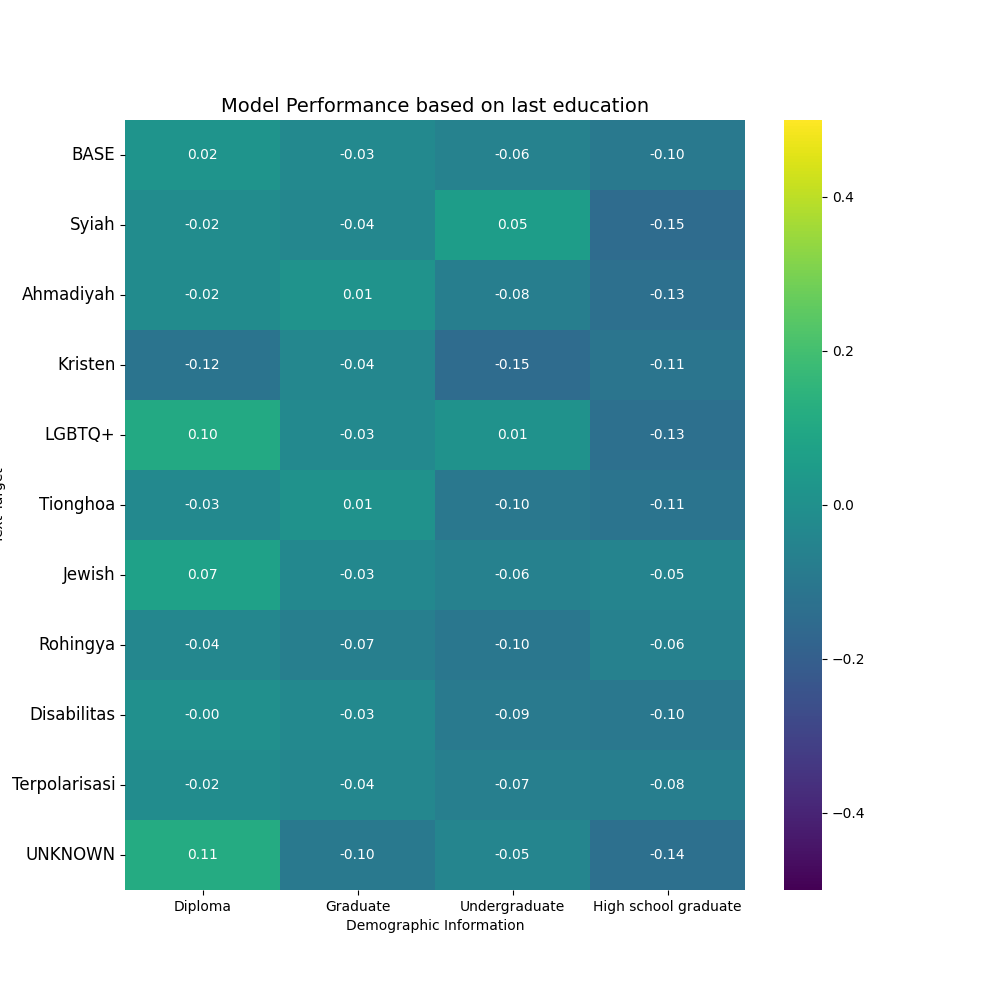}
\end{figure}
\begin{figure}
    \centering
    \includegraphics[width=1\textwidth]{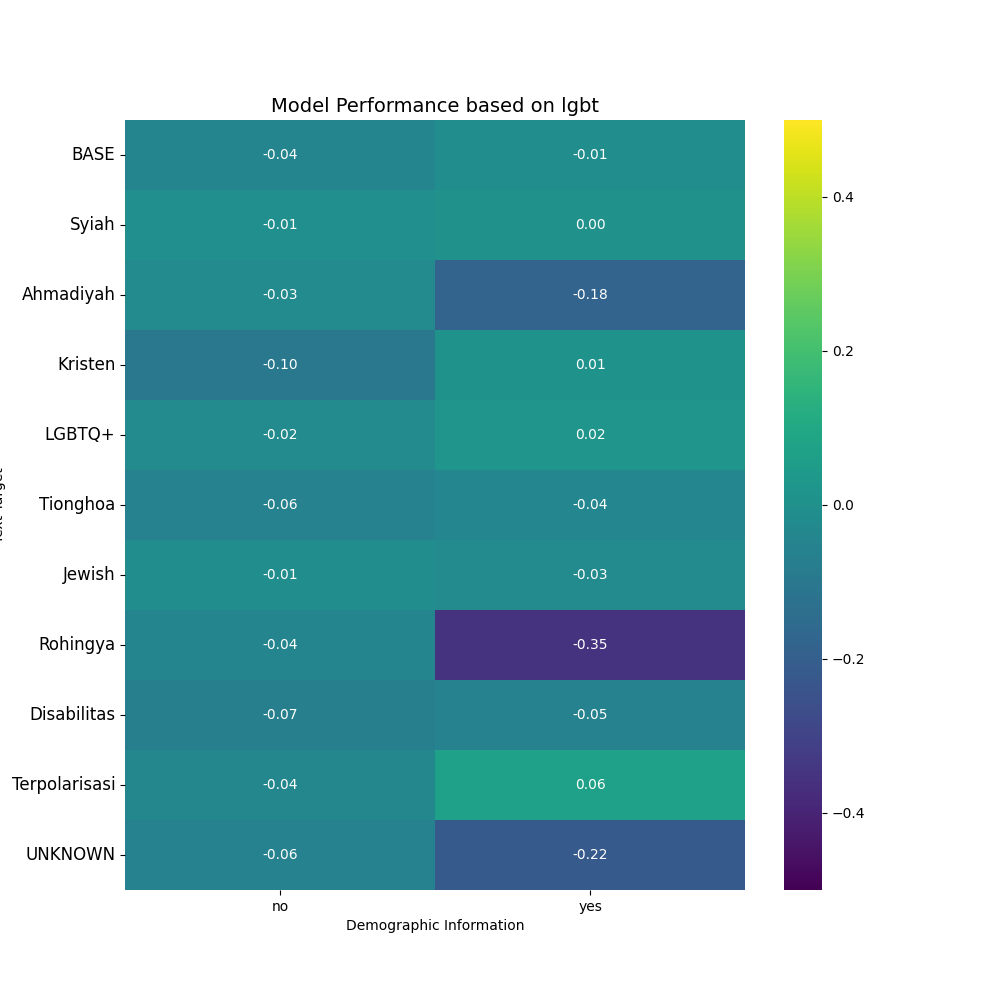}
\end{figure}
\begin{figure}
    \centering
    \includegraphics[width=1\textwidth]{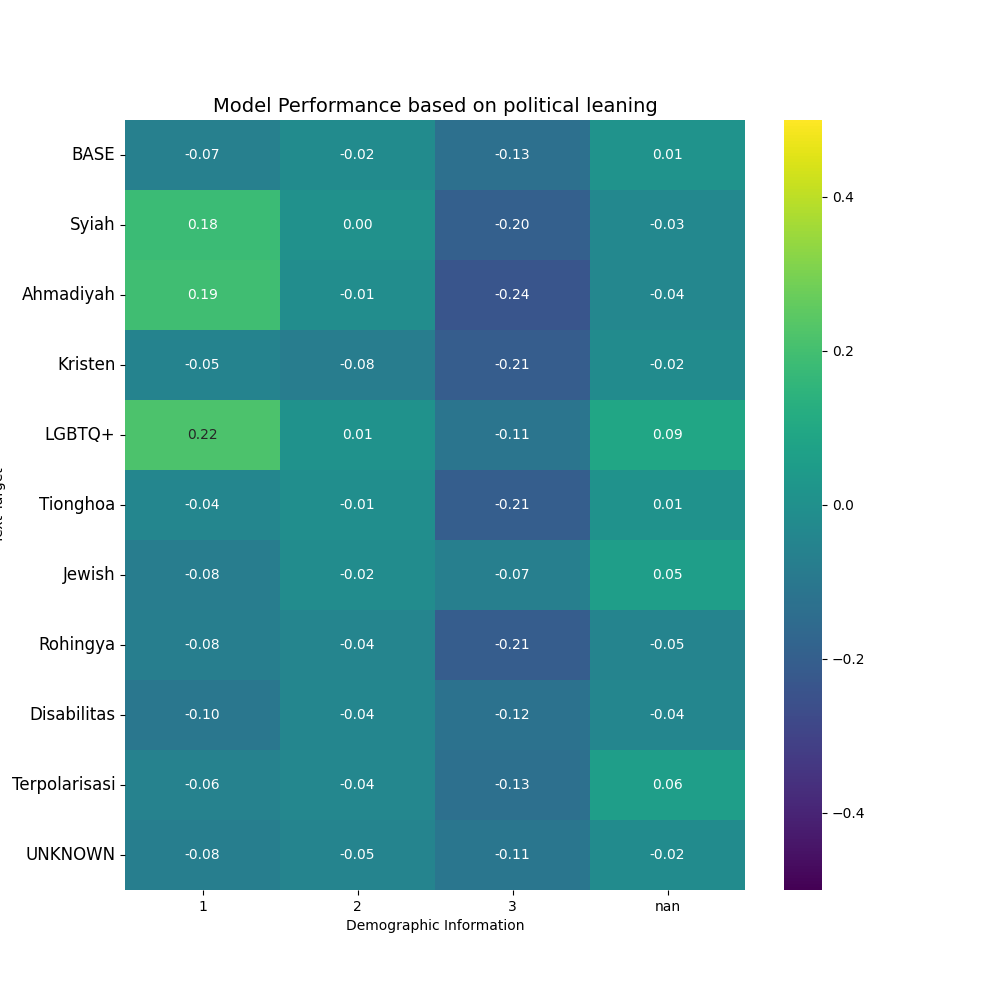}
\end{figure}
\begin{figure}
    \centering
    \includegraphics[width=1\textwidth]{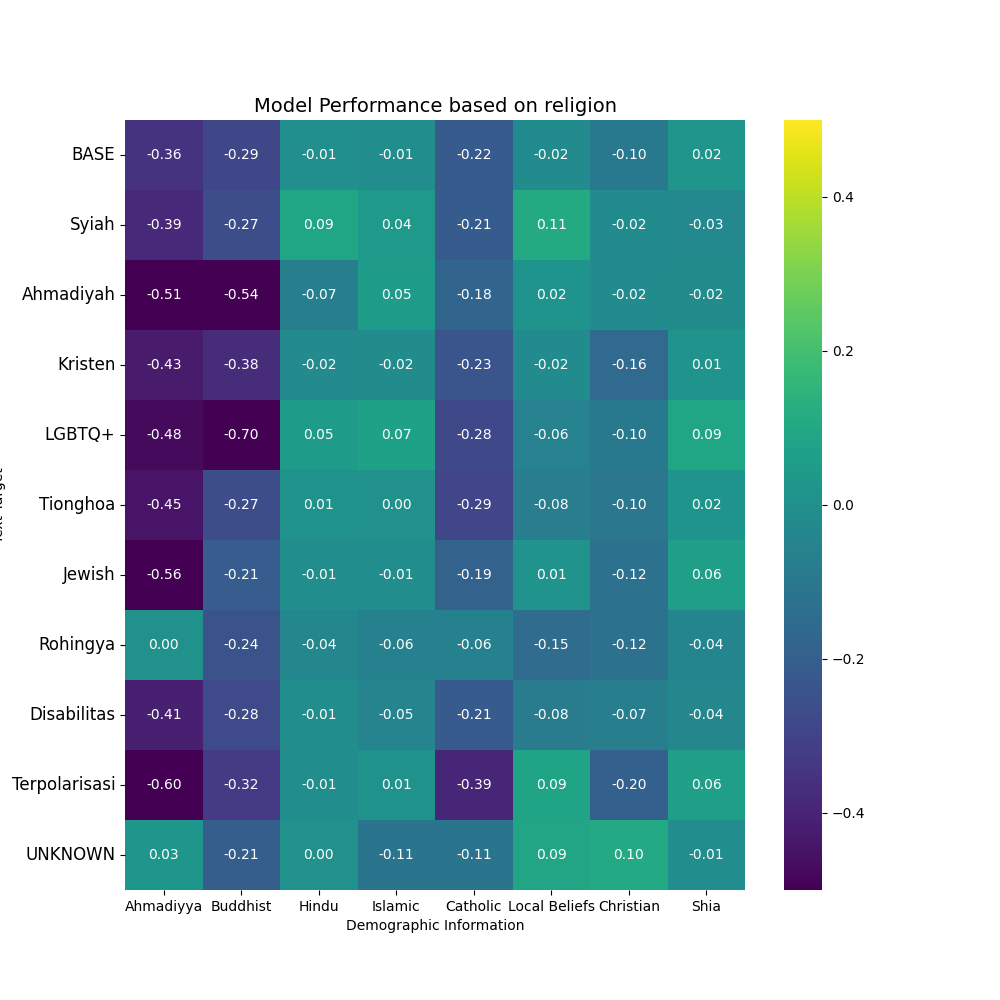}
\end{figure}

\newpage
\subsection{IndoBERTweet with Topic}
\begin{figure}[H]
    \centering
    \includegraphics[width=1\textwidth]{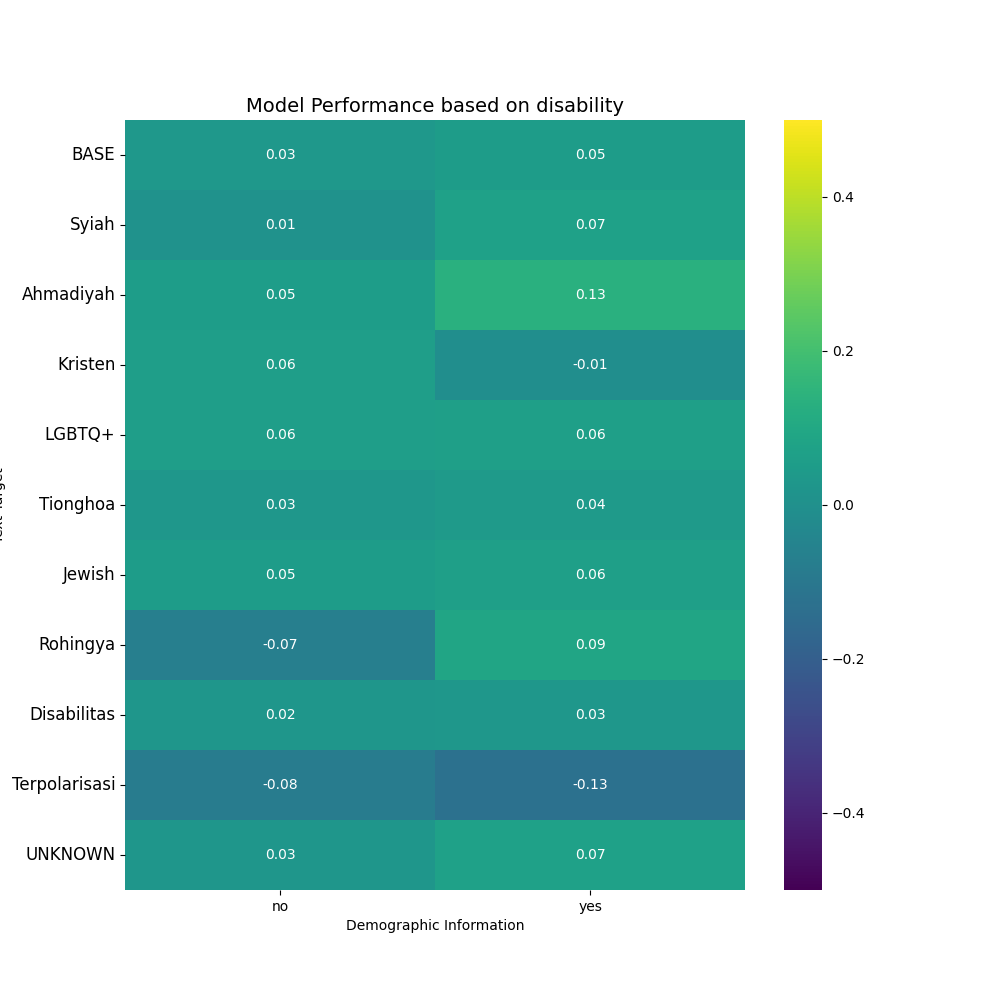}
\end{figure}
\begin{figure}
    \centering
    \includegraphics[width=1\textwidth]{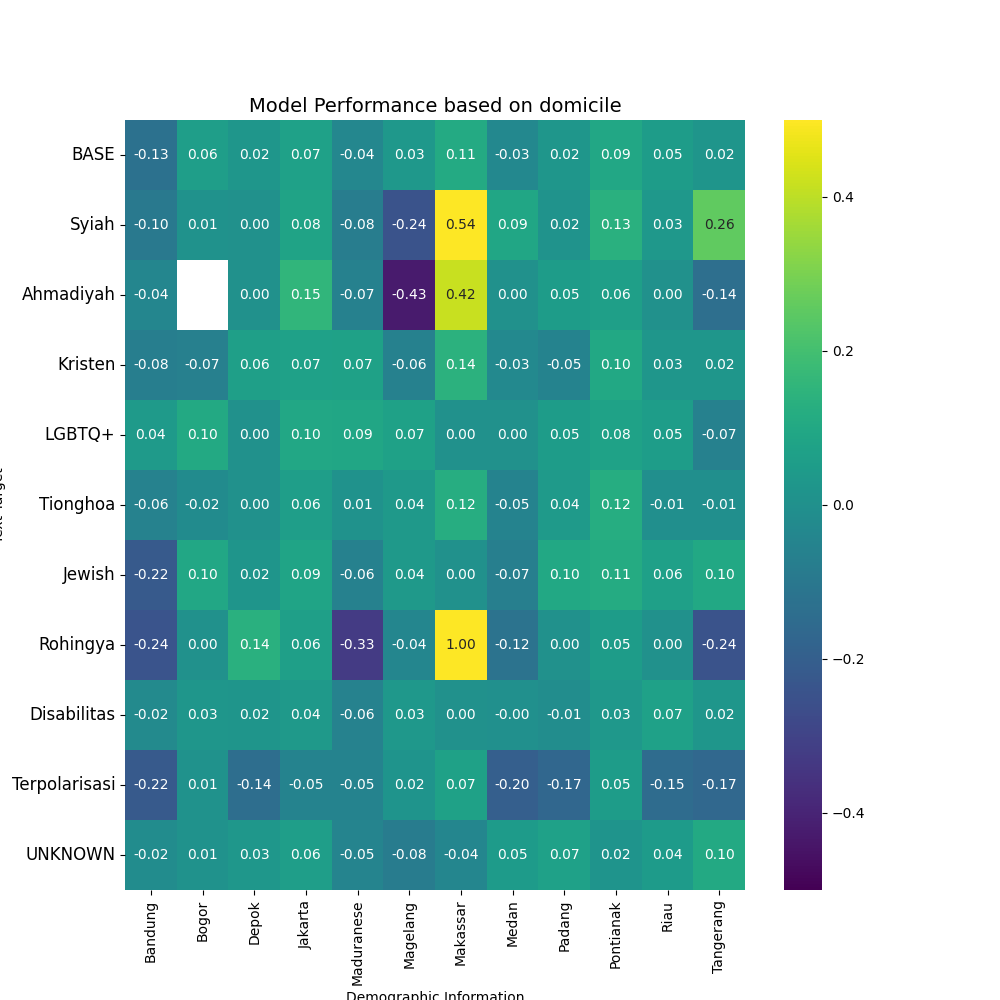}
\end{figure}
\begin{figure}
    \centering
    \includegraphics[width=1\textwidth]{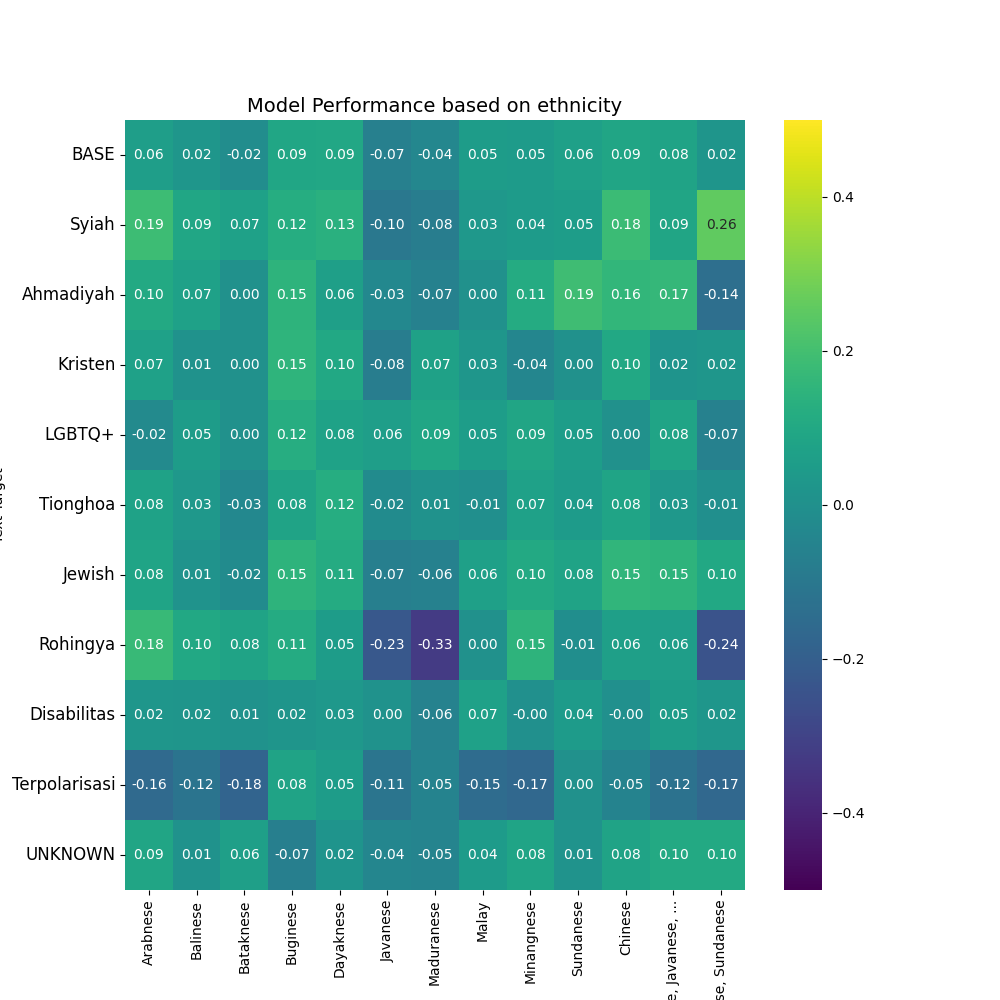}
\end{figure}
\begin{figure}
    \centering
    \includegraphics[width=1\textwidth]{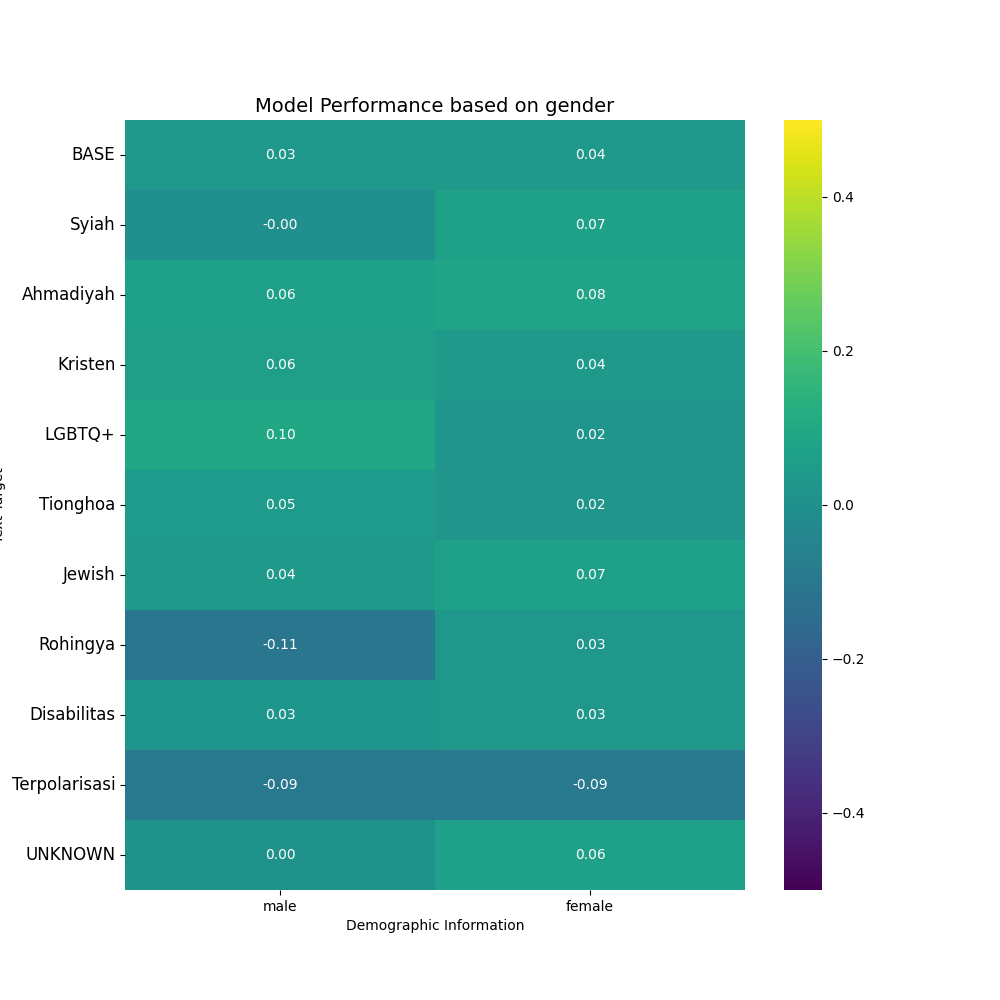}
\end{figure}
\begin{figure}
    \centering
    \includegraphics[width=1\textwidth]{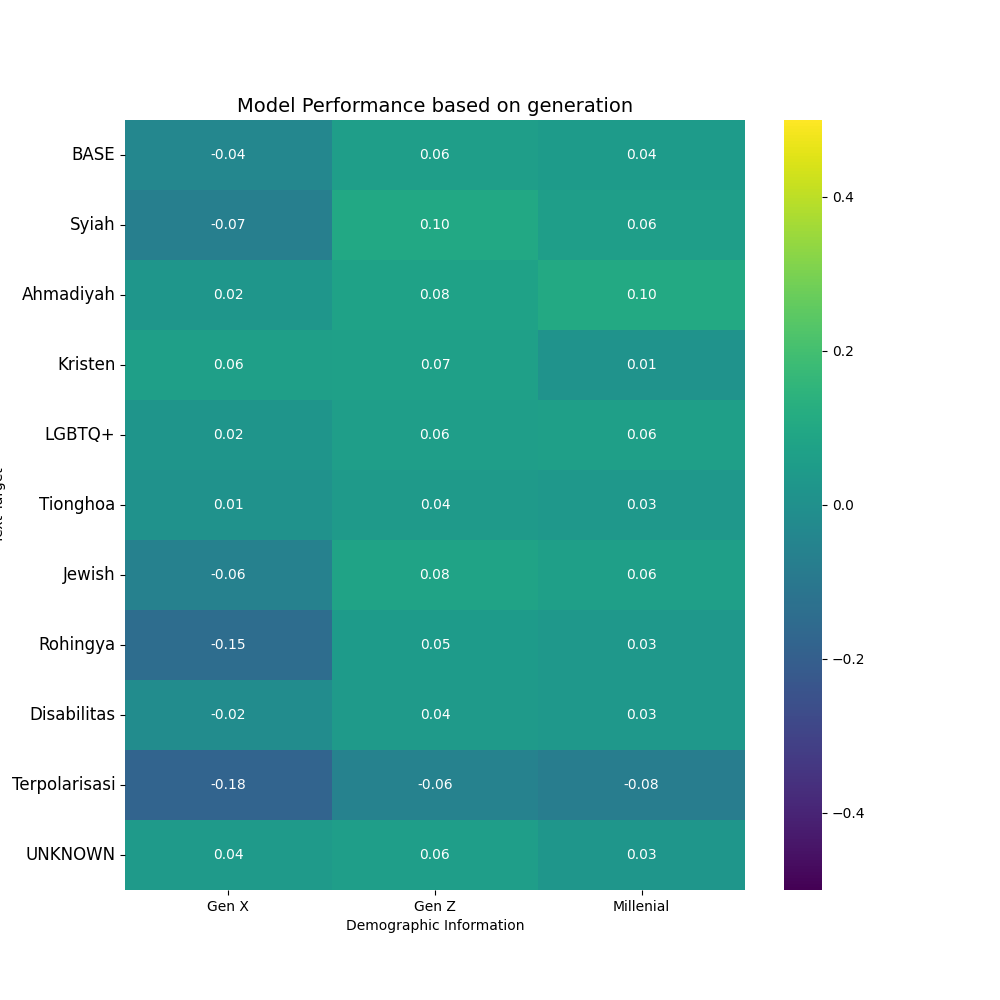}
\end{figure}
\begin{figure}
    \centering
    \includegraphics[width=1\textwidth]{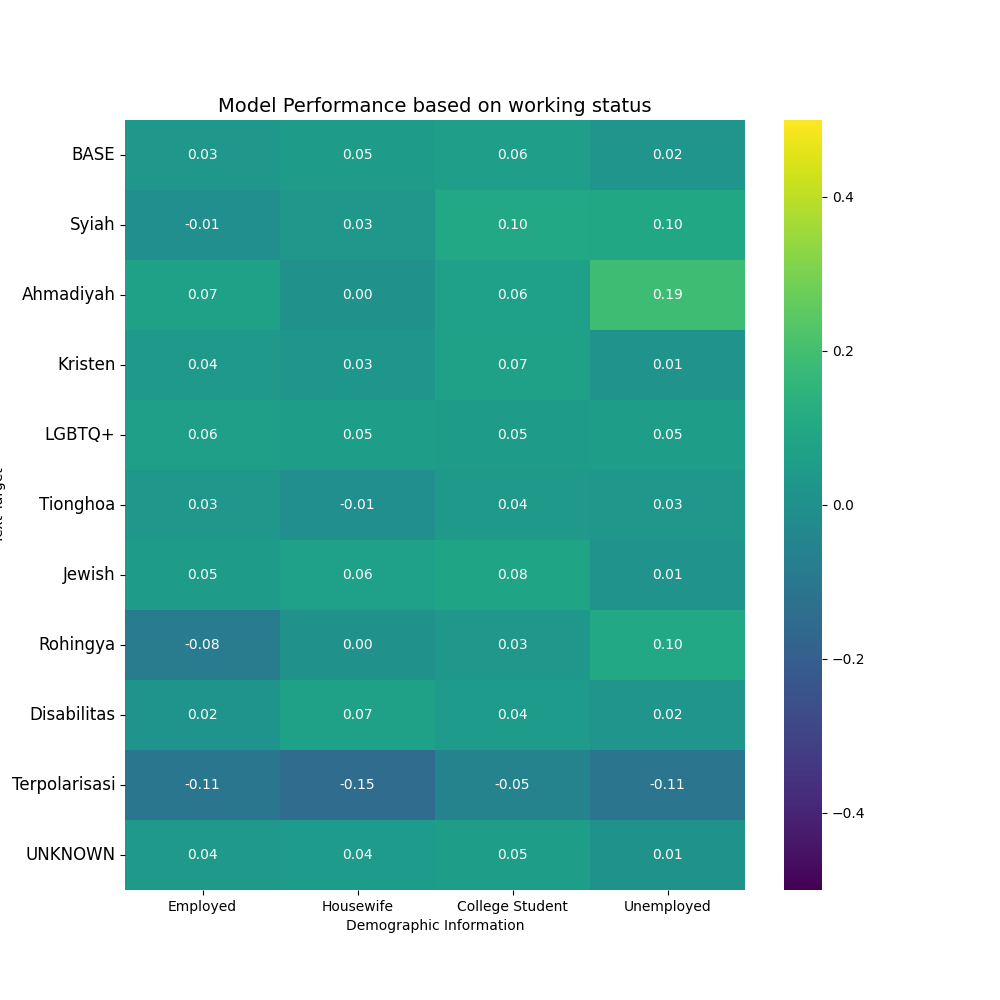}
\end{figure}
\begin{figure}
    \centering
    \includegraphics[width=1\textwidth]{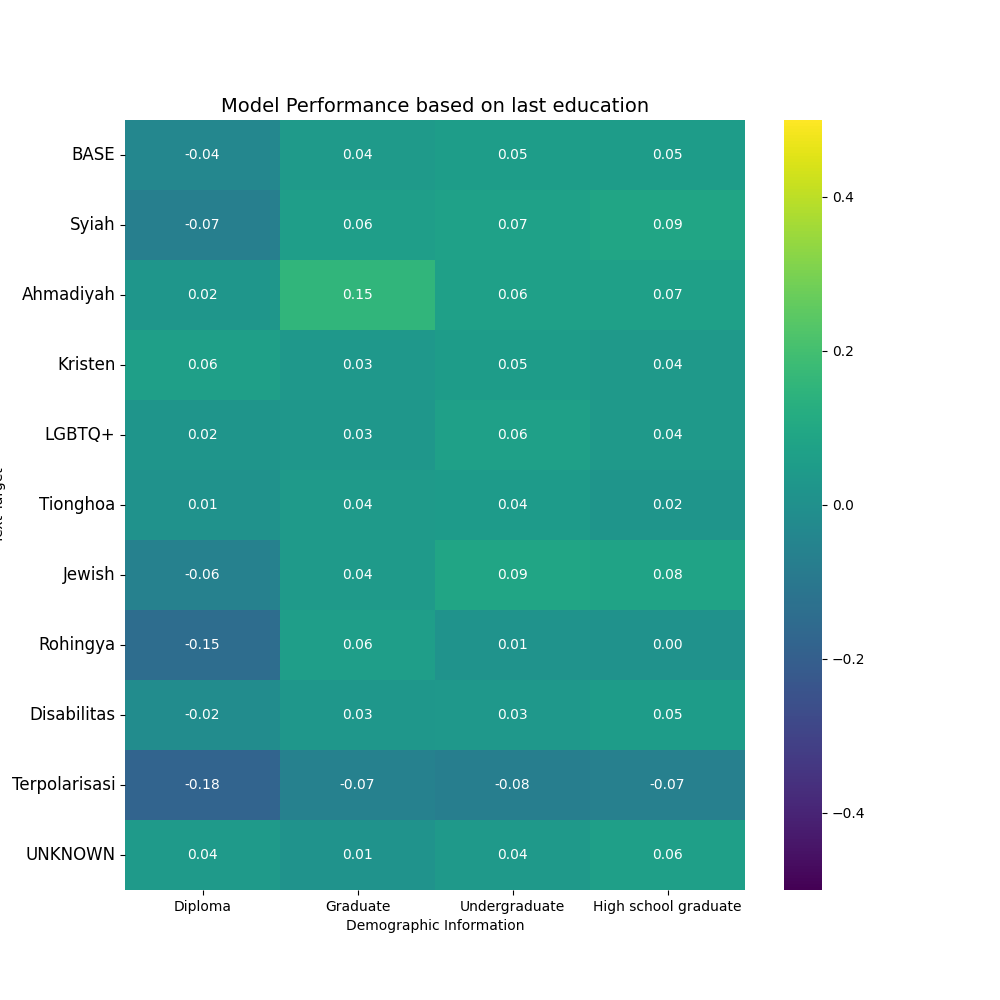}
\end{figure}
\begin{figure}
    \centering
    \includegraphics[width=1\textwidth]{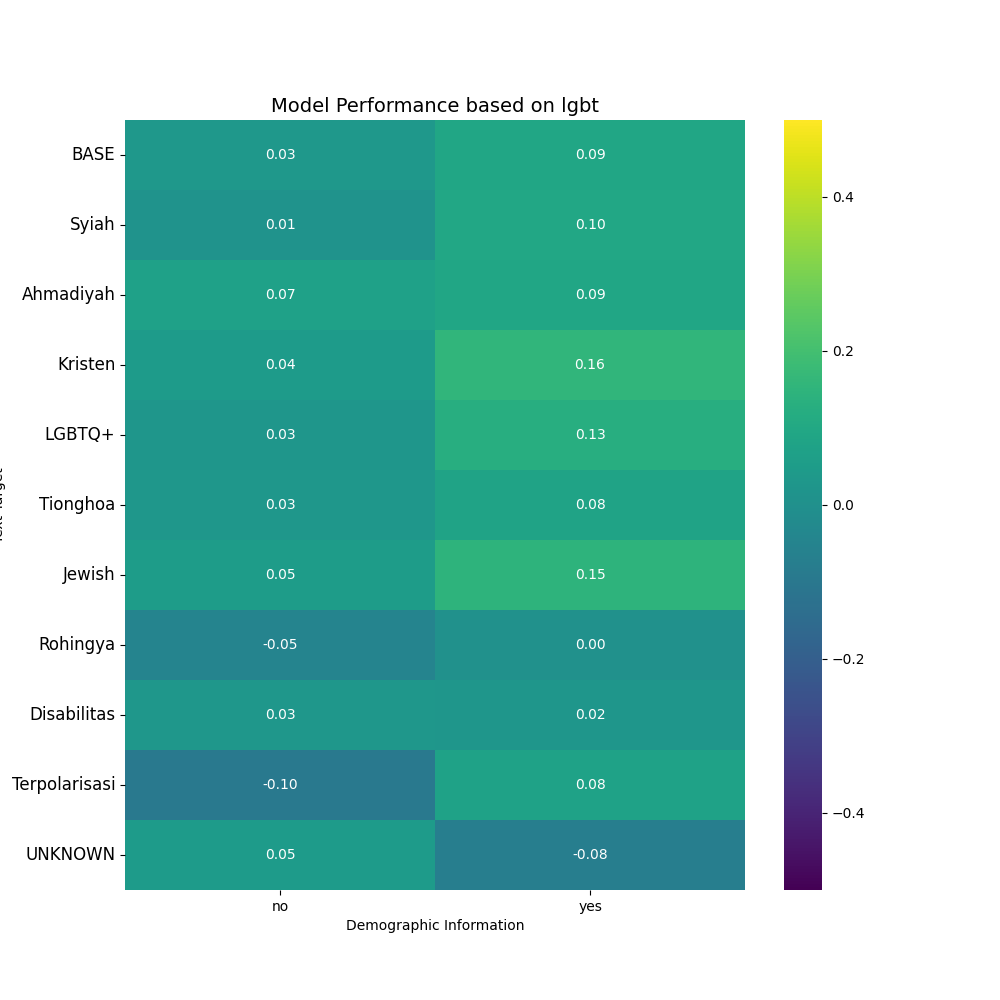}
\end{figure}
\begin{figure}
    \centering
    \includegraphics[width=1\textwidth]{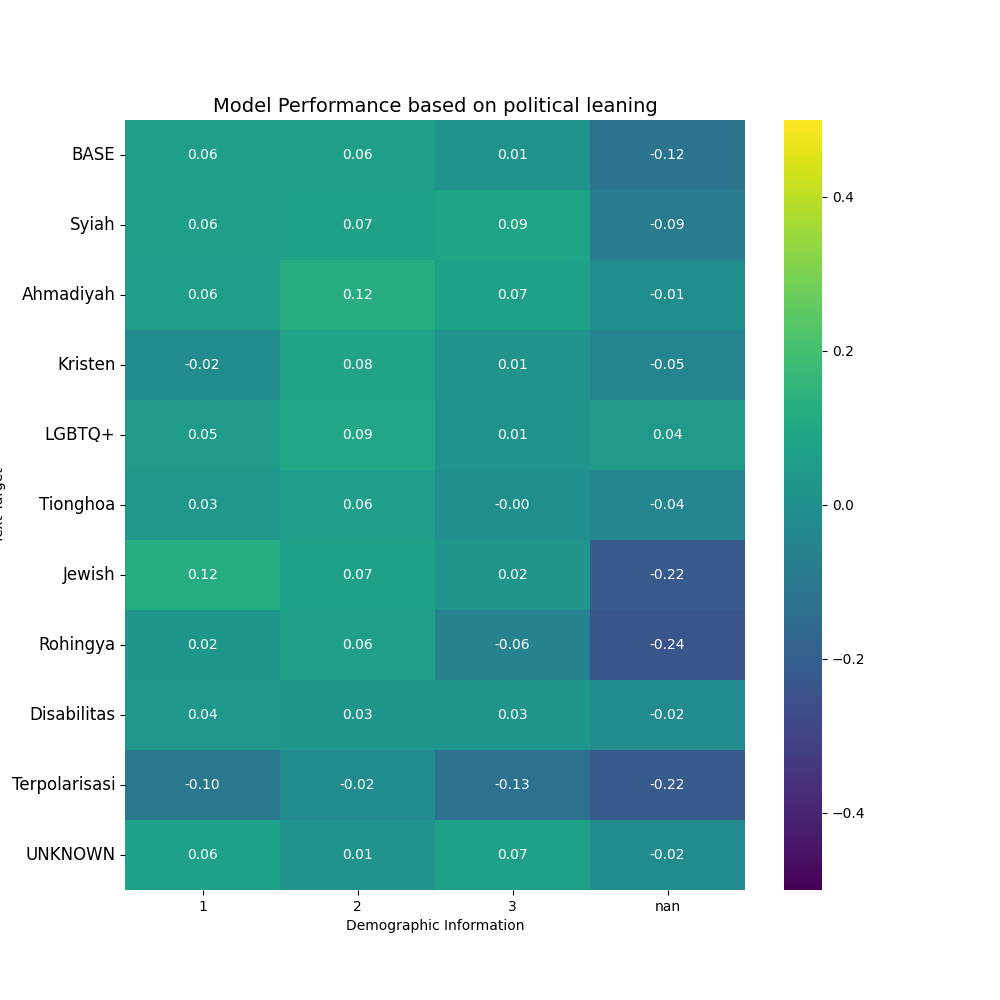}
\end{figure}
\begin{figure}
    \centering
    \includegraphics[width=1\textwidth]{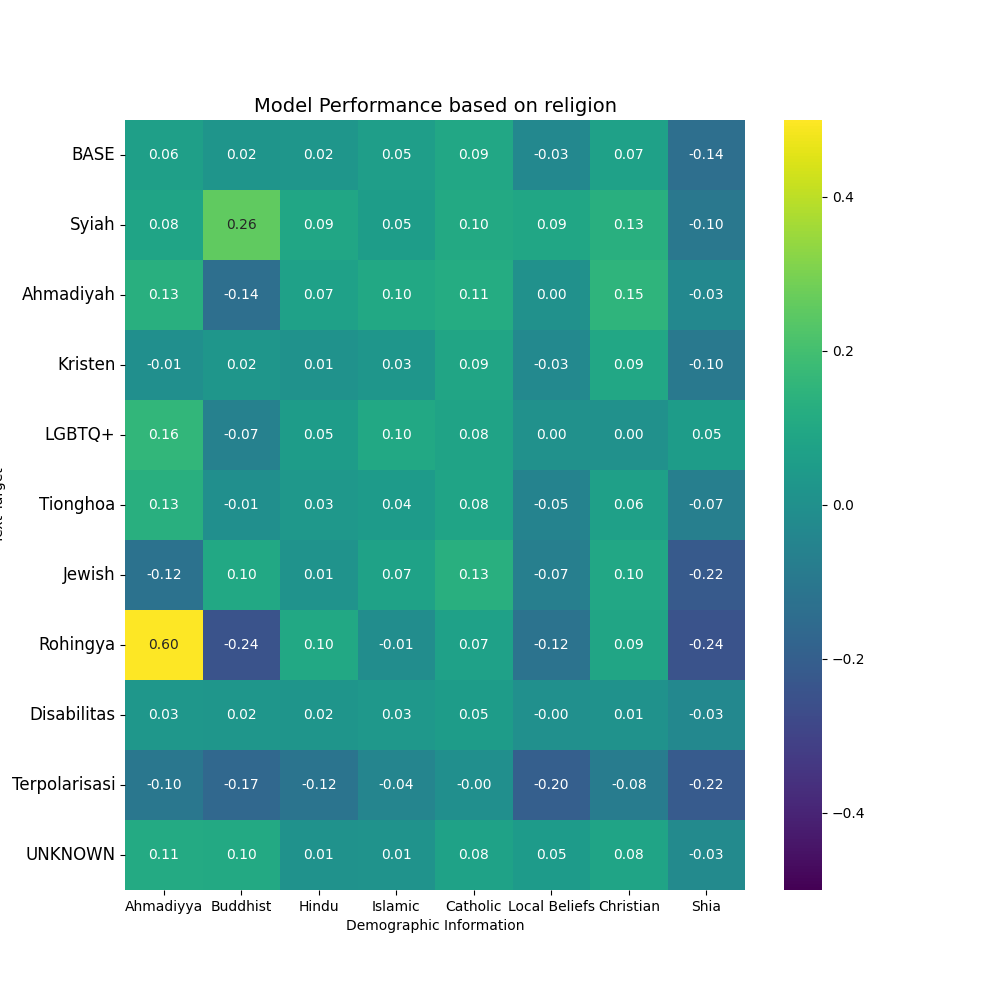}
\end{figure}

\newpage
\subsection{IndoBERTweet with Demographic and Topic}
\begin{figure}[H]
    \centering
    \includegraphics[width=1\textwidth]{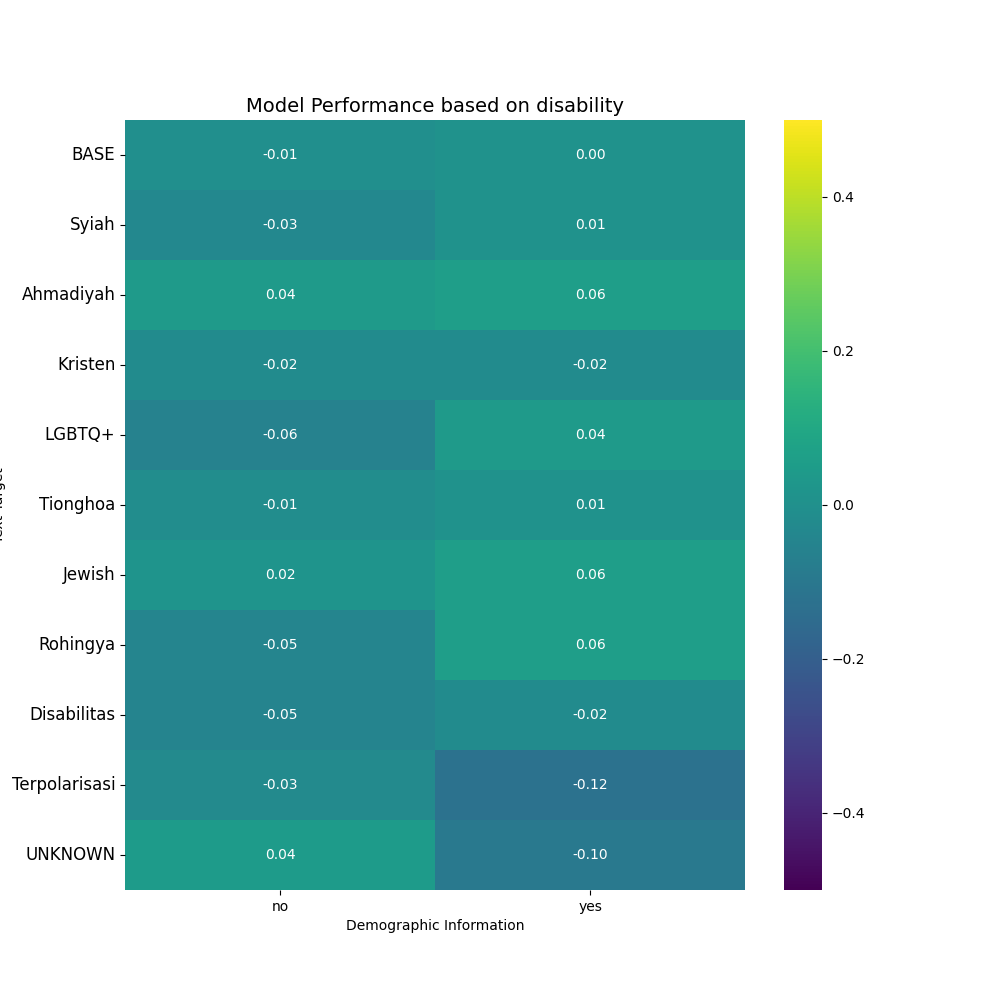}
\end{figure}
\begin{figure}
    \centering
    \includegraphics[width=1\textwidth]{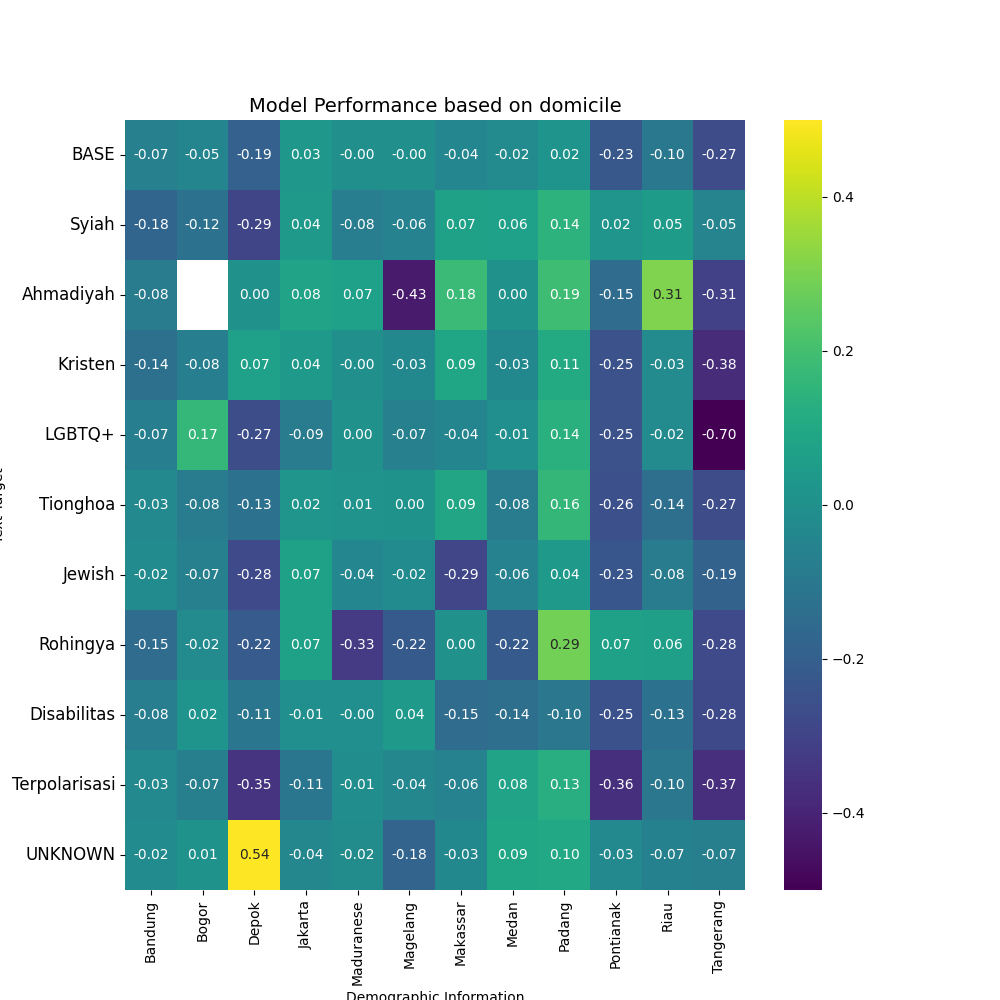}
\end{figure}
\begin{figure}
    \centering
    \includegraphics[width=1\textwidth]{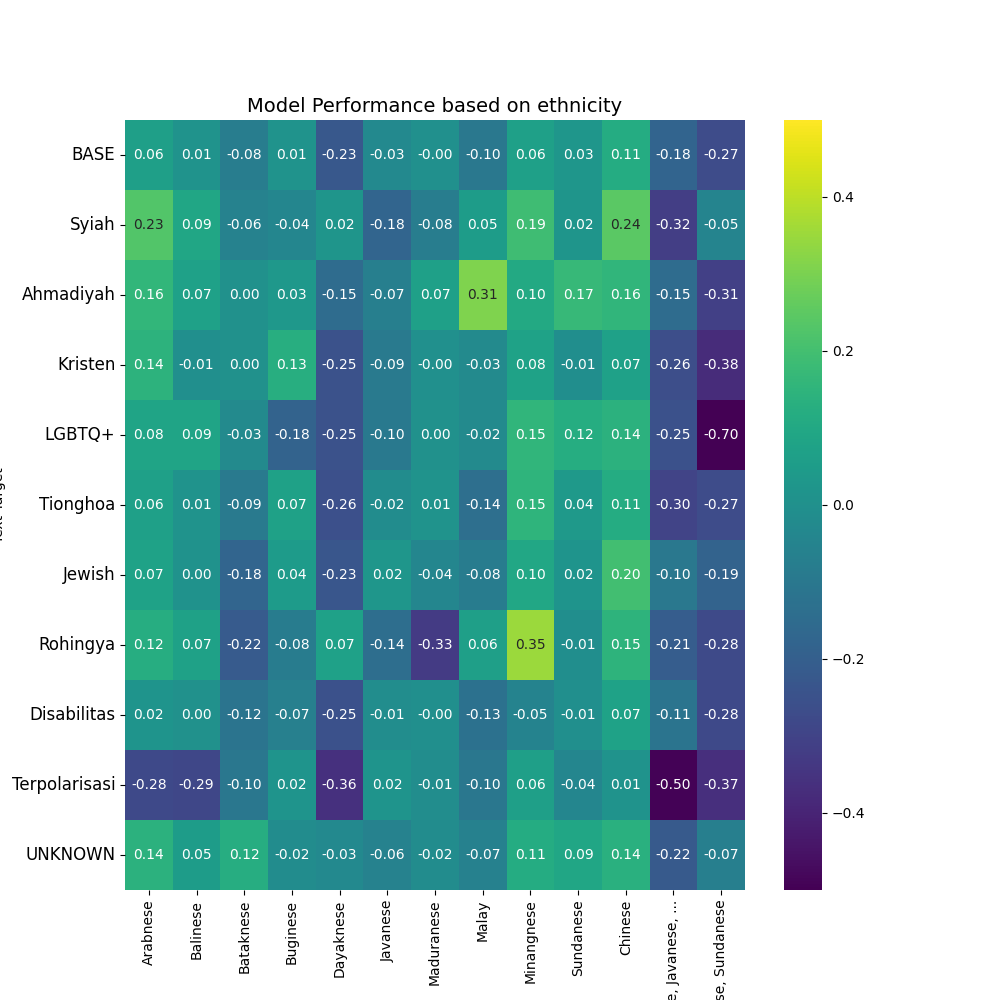}
\end{figure}
\begin{figure}
    \centering
    \includegraphics[width=1\textwidth]{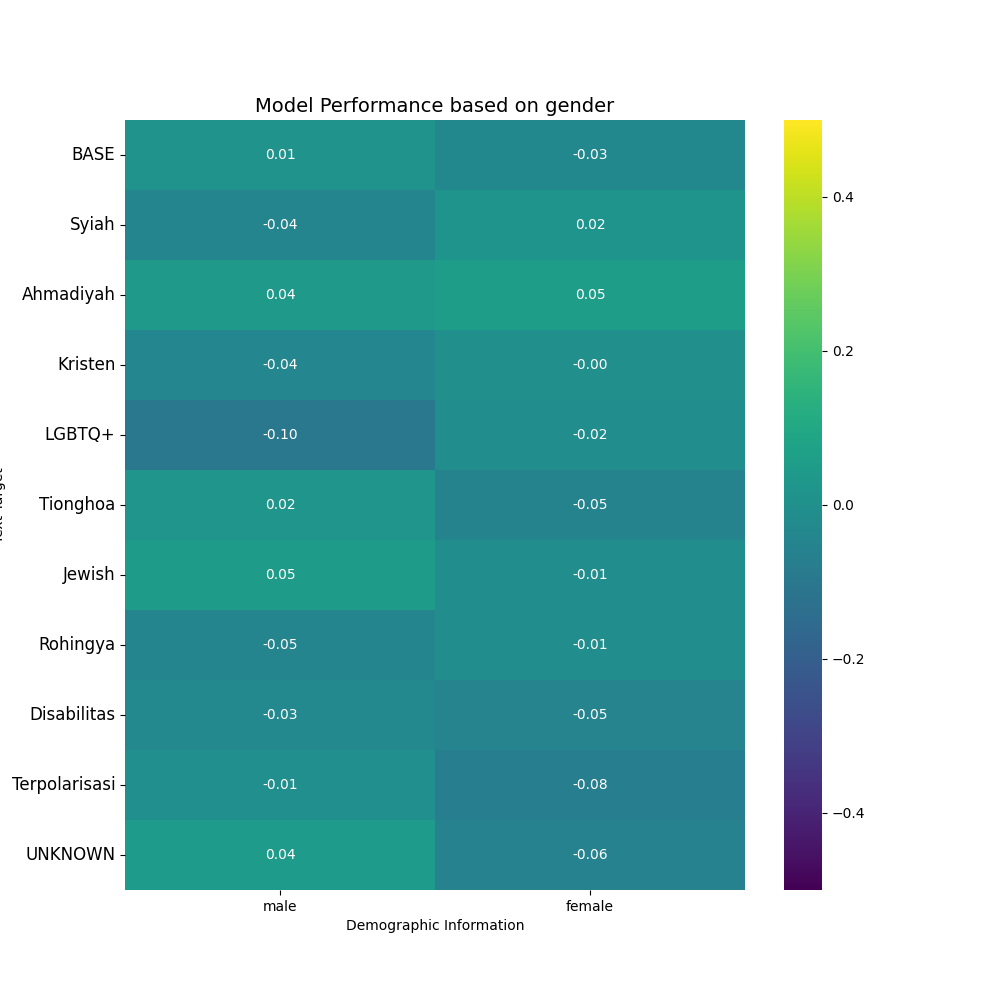}
\end{figure}
\begin{figure}
    \centering
    \includegraphics[width=1\textwidth]{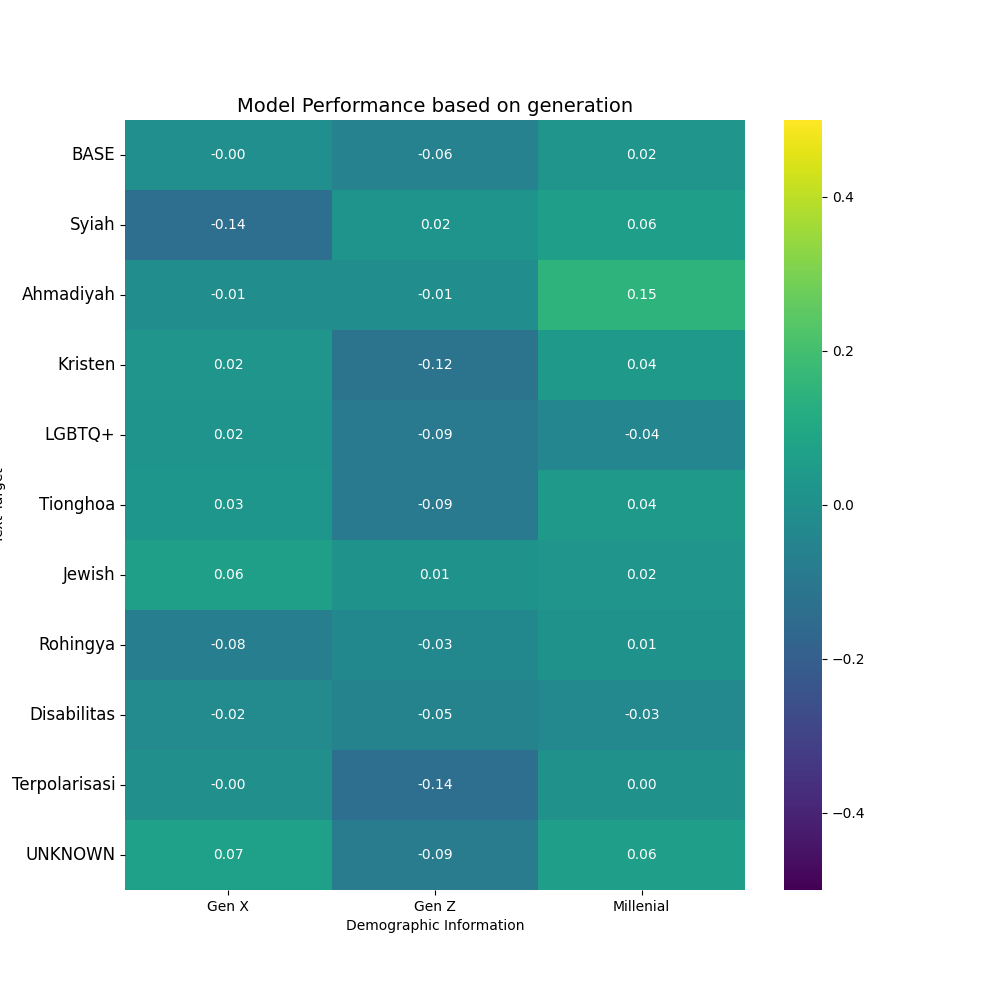}
\end{figure}
\begin{figure}
    \centering
    \includegraphics[width=1\textwidth]{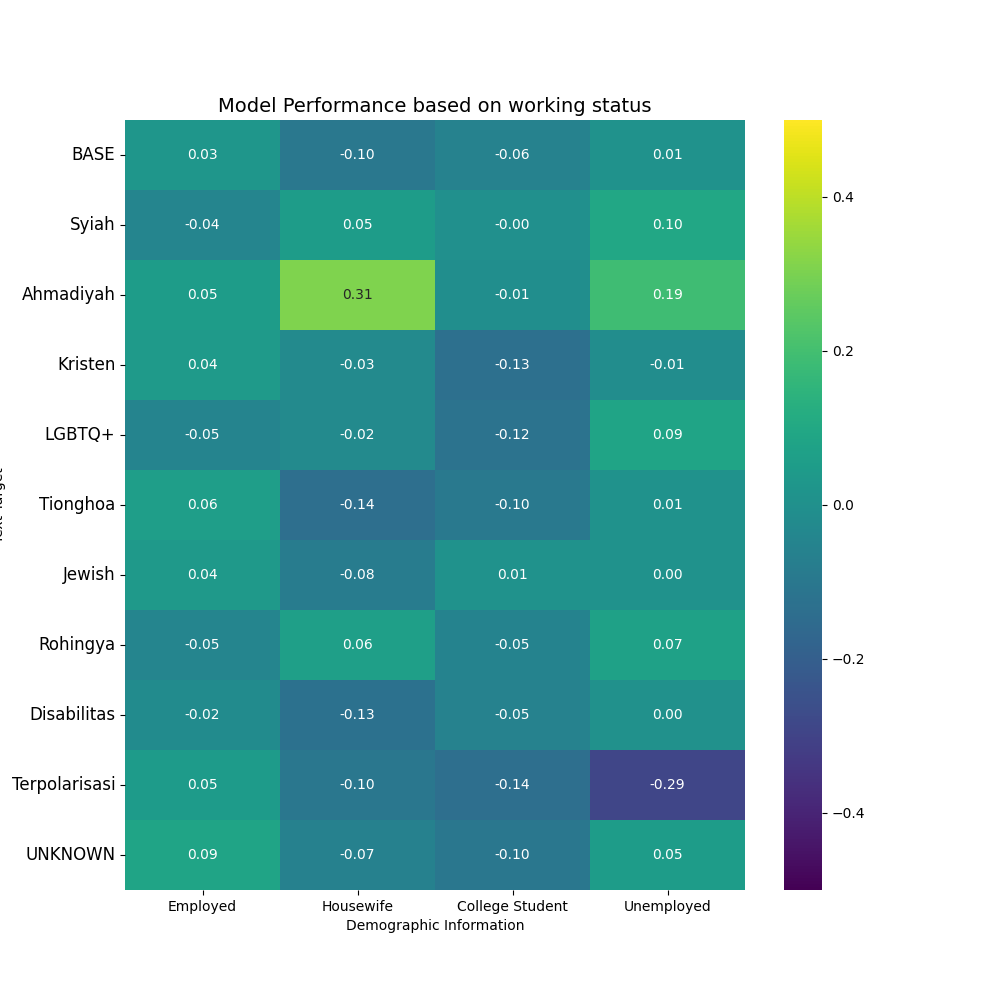}
\end{figure}
\begin{figure}
    \centering
    \includegraphics[width=1\textwidth]{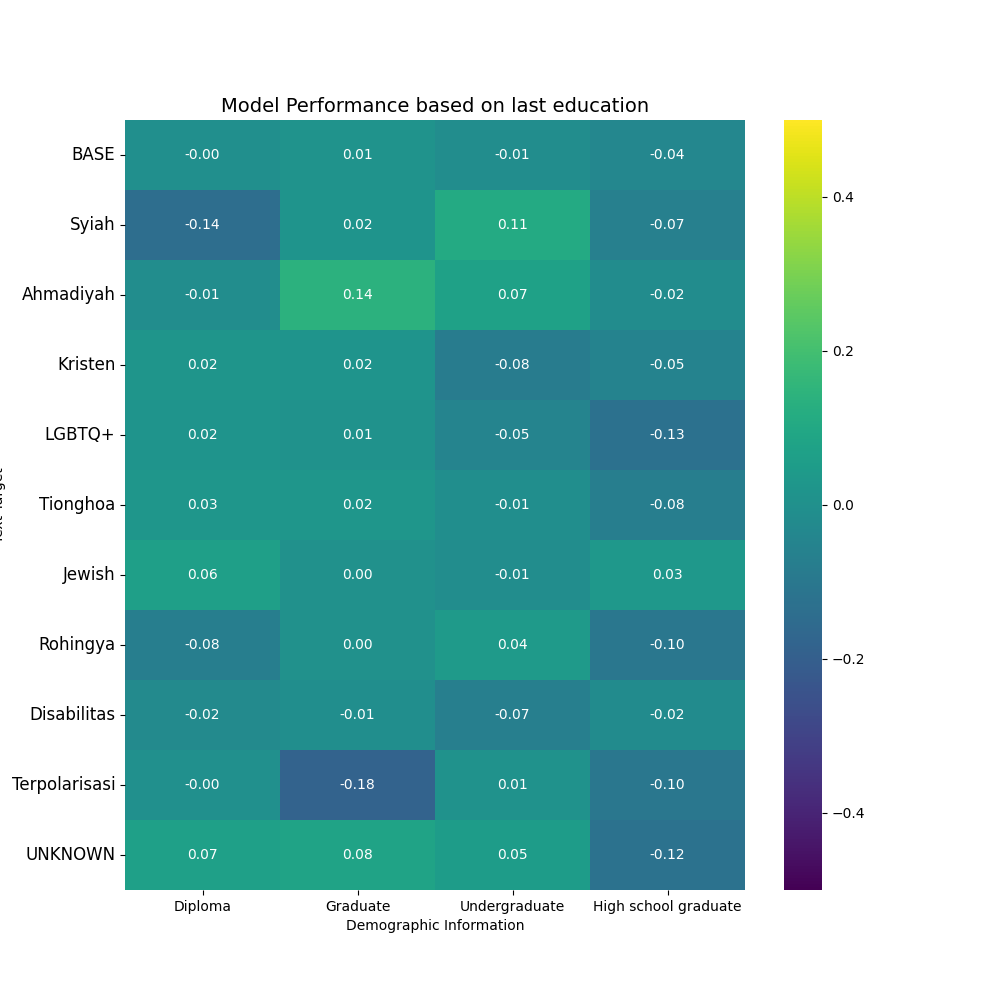}
\end{figure}
\begin{figure}
    \centering
    \includegraphics[width=1\textwidth]{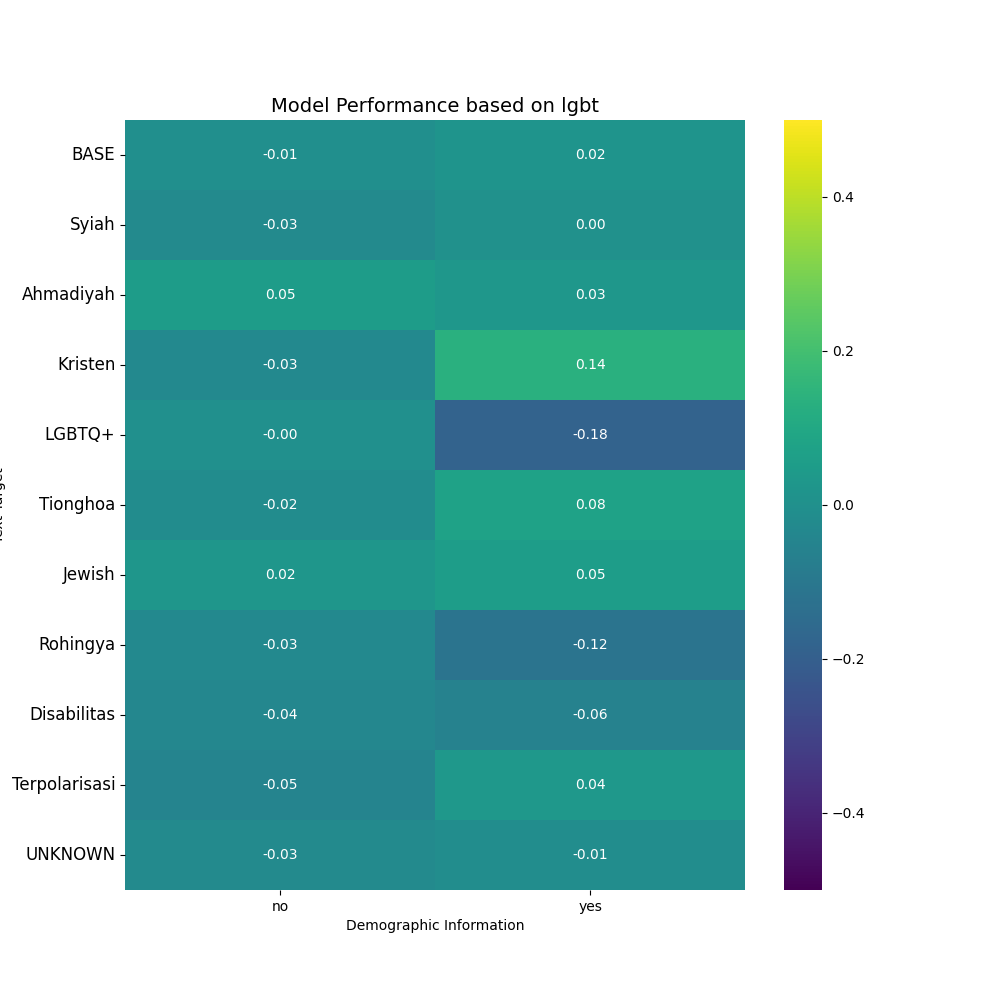}
\end{figure}
\begin{figure}
    \centering
    \includegraphics[width=1\textwidth]{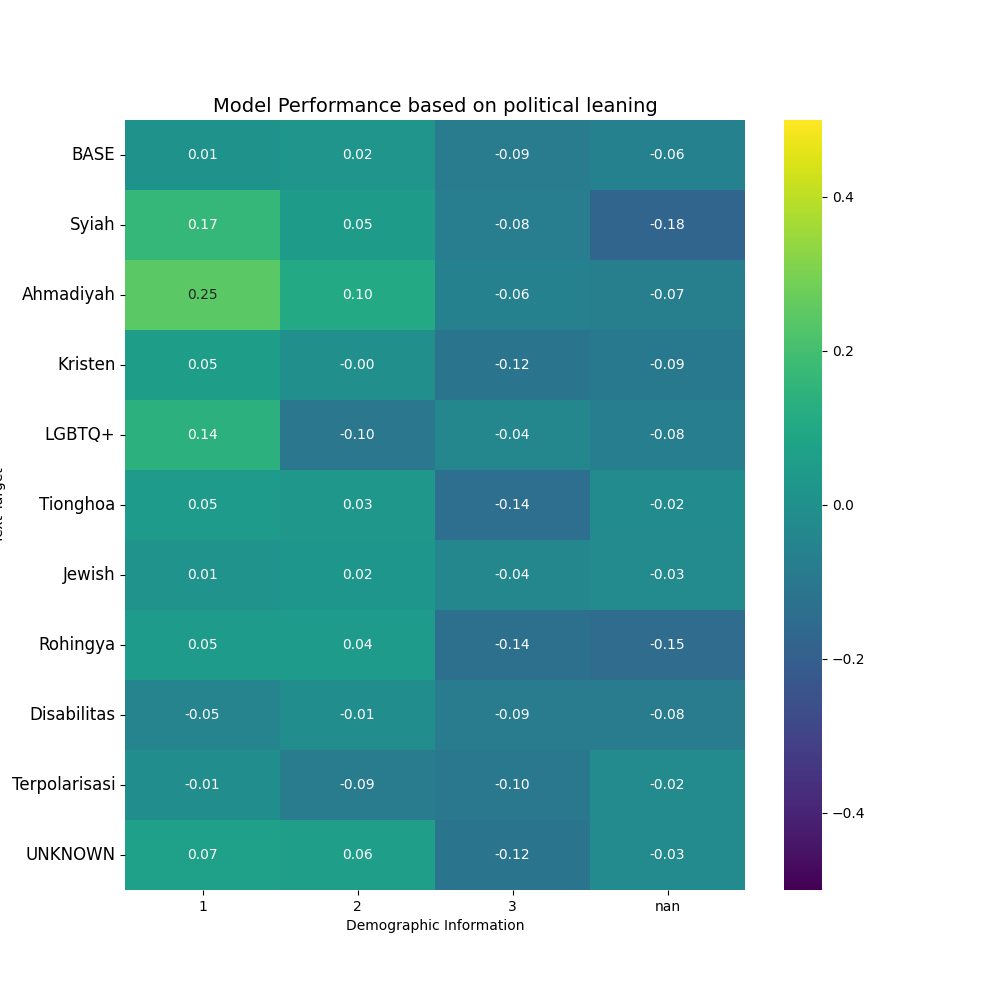}
\end{figure}
\begin{figure}
    \centering
    \includegraphics[width=1\textwidth]{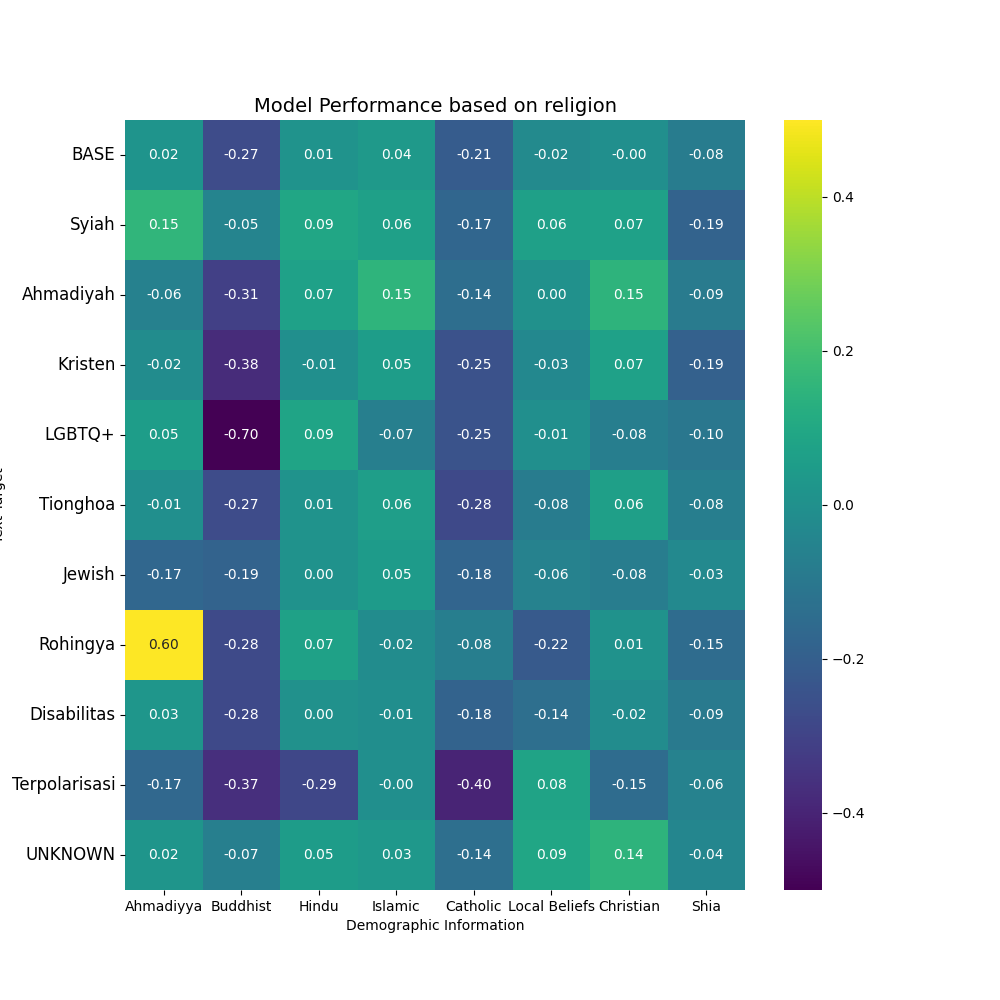}
\end{figure}
\end{document}